\documentclass[10pt,twocolumn,letterpaper]{article}

\usepackage{authblk}
\usepackage{cvpr}              % To produce the CAMERA-READY version

\usepackage[accsupp]{axessibility} % Improves PDF readability for those with visual impairments.
\usepackage[dvipsnames]{xcolor}

\usepackage[group-separator={,}]{siunitx}
\definecolor{cvprblue}{rgb}{0.21,0.49,0.74}
\usepackage[pagebackref,breaklinks,colorlinks,citecolor=cvprblue]{hyperref}
\usepackage{graphbox,graphicx}
\newcommand*\samethanks[1][\value{footnote}]{\footnotemark[#1]}

\title{Rich Human Feedback for Text-to-Image Generation}

\author[1]{Youwei Liang\thanks{Co-first authors, equal technical contribution}\thanks{The work was done during an internship at Google.\\Email: {\tt youwei@ucsd.edu}}}
\author[2]{Junfeng He\samethanks[1]\thanks{Corresponding authors, equal leading contribution.\\Email: {\tt\{junfenghe,leebird\}@google.com}}}
\author[2]{Gang Li\samethanks[1]\samethanks}
\author[5]{Peizhao Li\samethanks[2]}
\author[2]{Arseniy Klimovskiy}
\author[2]{Nicholas Carolan}
\author[3]{Jiao Sun\samethanks[2]\thanks{Currently affiliated with Google Gemini Team}}
\author[2]{Jordi Pont-Tuset}
\author[2]{Sarah Young}
\author[2]{Feng Yang}
\author[2]{Junjie Ke}
\author[2]{Krishnamurthy Dj Dvijotham}
\author[4]{Katherine M. Collins\samethanks[2]}
\author[2]{Yiwen Luo}
\author[2]{Yang Li}
\author[2]{Kai J Kohlhoff}
\author[2]{Deepak Ramachandran}
\author[2]{Vidhya Navalpakkam}

\affil[1]{University of California, San Diego}
\affil[2]{Google Research}
\affil[3]{University of Southern California}
\affil[4]{University of Cambridge}
\affil[5]{Brandeis University}

\begin{document}
\maketitle

\begin{abstract}
Recent Text-to-Image (T2I) generation models such as Stable Diffusion and Imagen have made significant progress in generating high-resolution images based on text descriptions. However, many generated images still suffer from issues such as artifacts/implausibility, misalignment with text descriptions, and low aesthetic quality. 
Inspired by the success of Reinforcement Learning with Human Feedback (RLHF) for large language models, prior works collected human-provided scores as feedback on generated images and trained a reward model to improve the T2I generation. 
In this paper, we enrich the feedback signal by (i) marking image regions that are implausible or misaligned with the text, and (ii) annotating which words in the text prompt are misrepresented or missing on the image. 
We collect such rich human feedback on 18K generated images (RichHF-18K) and train a multimodal transformer to predict the rich feedback automatically. 
We show that the predicted rich human feedback can be leveraged to improve image generation, for example, by selecting high-quality training data to finetune and improve the generative models, or by creating masks with predicted heatmaps to inpaint the problematic regions. 
Notably, the improvements generalize to models (Muse) beyond those used to generate the images on which human feedback data were collected (Stable Diffusion variants). The RichHF-18K data set will be released in our GitHub repository: \href{https://github.com/google-research/google-research/tree/master/richhf_18k}{https://github.com/google-research/google-research/tree/master/richhf\_18k}.
\end{abstract}

%-------------------------------------------------------------------------

\section{Introduction}
% \TODO{Youwei and Junfeng: Let's pitch this paper as the first "rich human feedback" paper for image generation. Avoid being another or a better image reward paper.}

% \TODO{Please help rewrite/rephrase the whole paper to reduce space!
% }

% \TODO{Use the following names. 2 Heatmaps: implausibility heatmap, misalignment heatmap.
% 4 scores: plausibility score, (text-image) alignment score, aesthetics score, overall score.
% 1 sequence: misaligned keyword sequence
% }

\label{sec:intro}
Text-to-image (T2I) generation models~\cite{saharia2022imagen,croitoru2023diffusionSurvey,yang2022diffusionSurvey,zhang2023t2idiffusion,rombach2022stablediffusion,yu2022scaling,gu2023matryoshka} are rapidly becoming a key to content creation in various domains, including entertainment, art, design, and advertising, and are also being generalized to image editing~\cite{sheynin2023emu,brooks2023instructpix2pix,kawar2023imagic,wang2023imageneditor}, video generation~\cite{ho2022imagenVideo,Ni_2023_CVPR,xing2023survey}, among many other applications. 
Despite significant recent advances, the outputs still usually suffer from issues such as artifacts/implausibility, misalignment with text descriptions, and low aesthetic  quality~\cite{xu2023imagereward,kirstain2023pickapic,wu2023hps}. 
% In the ideal T2I generation, the generated images should be free of artifacts (\ie, any unnatural patterns, spots, or distortions), and should be plausible (meaning anything on the image should look like what it typically appears in real life) and well aligned with the text prompts. 
% % The text-image alignment requirement includes a wide variety of aspects such as object categories, colors, spatial positions, and actions. 
% Last but not least, the generated image should also follow the aesthetic, emotional, and sensational requirements in the text prompts. 
% Unfortunately, the state-of-the-art (SOTA) text-to-image generation models fail to satisfy all the requirements.
For example, in the Pick-a-Pic dataset~\cite{kirstain2023pickapic}, which mainly consists of images generated by Stable Diffusion model variants
% and Dreamlike Photoreal models~\cite{rombach2022stablediffusion}
, many images (\eg{} \cref{fig:annotation}) contain distorted human/animal bodies (\eg{} human hands with more than five fingers), distorted objects and implausibility issues such as a floating lamp.
Our human evaluation experiments find that only $\sim$10\% of the generated images in the dataset are free of artifacts and implausibility.
Similarly, text-image misalignment issues are common too, \eg{}, the prompt is ``\textit{a man jumping into a river}'' but the generated image shows the man standing.
% Moreover, sometimes the generated images are misaligned with the text prompt in some of the keywords in the prompt, either misrepresenting the keywords (\ie, wrong colors / spatial relations) or having the required objects completely missing~\cite{kirstain2023pickapic}. 

Existing automatic evaluation metrics for generated images, however, including the well-known IS~\cite{salimans2016IS} and %Fr\'{e}chet Inception Distance 
FID~\cite{heusel2017FID}, are computed over distributions of images and may not reflect nuances in individual images.
Recent research has collected human preferences/ratings to evaluate the quality of generated images and trained evaluation models to predict those ratings~\cite{xu2023imagereward, kirstain2023pickapic,wu2023hps}, notably ImageReward~\cite{xu2023imagereward} or Pick-a-Pic~\cite{kirstain2023pickapic}. 
While more focused, these metrics still summarize the quality of one image into a single numeric score.
In terms of prompt-image alignment, there are also seminal single-score metrics such as CLIPScore~\cite{jack-hessel-2021} and more recent question generation and answering pipelines~\cite{yushi-hu-2023,yarom2023read,jaemin-cho-2023,JaeminCho2023}.
While more calibrated and explainable, these are expensive and complex models that still do not localize the regions of misalignment in the image.

In this paper, we propose a dataset and a model of fine-grained multi-faceted evaluations that are interpretable and attributable (\eg{}, to regions with artifacts/implausibility or image-text misalignments), which provide a much richer understanding of the image quality than single scalar scores.
% In such a background, it is very helpful to collect rich human feedback to evaluate generated images reliably, with not only one overall score score, but also more feedback as rationale and insights about what which part went wrong, etc. 
% Moreover, an automatic evaluation method for text-to-image generation can also be used as a reward model or auxiliary objective function to improve the generative models. 
% Although it is difficult to collect user-edited images for T2I generation, it is possible to collect rich human feedback for generated images. Rich human feedback is helpful in improving T2I generation as it can specify which parts of the images are implausible or misaligned with text prompts, which is also more explainable when evaluating T2I models. However, existing works have mainly focused on collecting human scores or rankings on generated images. 
%However, all these works only collect human scores on the generated images and have not yet collected any fine-grained feedback. 
As a first contribution, we collect a dataset of Rich Human Feedback on 18K images (RichHF-18K), which contains (i) point annotations on the image that highlight regions of implausibility/artifacts, and text-image misalignment; (ii) labeled words on the prompts specifying the missing or misrepresented concepts in the generated image; and (iii) four types of fine-grained scores for image plausibility, text-image alignment, aesthetics, and overall rating.
Equipped with RichHF-18K, we design a multimodal transformer model, which we coin as Rich Automatic Human Feedback (RAHF) to learn to predict these rich human annotations on generated images and their associated text prompt.
Our model can therefore predict implausibility and misalignment regions, misaligned keywords, as well as fine-grained scores.
This not only provides reliable ratings, but also more detailed and explainable insights about the quality of the generated images.
% We leverage a Vision Transformer (ViT)~\cite{vit,maxvit} encoder to encode the image and a text transformer (T5X~\cite{t5x}) to encode the text prompt. Then we utilize self-attention modules to fuse the image and text information using the embedding from the ViT and the T5X. On top of the self-attention modules, an MLP head produces fine-grained scores, a convolutional decoder generates heatmaps to predict the fine-grained point annotations, and a text decoder generates a text sequence to predict the misaligned keywords. 
% Our reward model is trained end-to-end with all the rich feedback data and exhibits generalization to images generated by various models. 
To the best of our knowledge, this is the first rich feedback dataset and model for state-of-the-art text-to-image generation models, providing an automatic and explainable pipeline to evaluate T2I generation. 

% \TODO{Name our dataset RichHF-18K, and use this name instead of Pick-a-Pic to refer to our dataset for the rest of our paper. A better name than RichHF-18K might be needed.}

\vspace{1mm}
\noindent The main contributions of this paper are summarized below:
\begin{enumerate}
\item The first Rich Human Feedback dataset (RichHF-18K) on generated images (consisting of fine-grained scores, implausibility(artifact)/misalignment image regions, and misalignment keywords), on 18K Pick-a-Pic images. 

\item A multimodal Transformer model (RAHF) to predict rich feedback on generated images, which we show to be highly correlated with the human annotations on a test set. %Moreover, we provide two model variants, a simple multi-head version, and moreover, an augmented prompt version that works better. 
% The model is named RAHF: Rich Automatic Human Feedback. 

\item We further demonstrate the usefulness of the predicted rich human feedback by RAHF to improve image generation: (i) by using the predicted heatmaps as masks to inpaint problematic image regions and (ii) by using the predicted scores to help finetune image generation models (like Muse~\cite{chang2023muse}), \eg, via selecting/filtering finetuning data, or as reward guidance. We show that in both cases we obtain better images than with the original model. 
\item The improvement on the Muse model, which differs from the models that generated the images in our training set, shows the good generalization capacity of our RAHF model.
%Even using a different generative model (Muse~\cite{chang2023muse}) than those in our training data RichHF-18K, the improvement on image generation can still be achieved, showing the good generalization capacity of our RAHF model.
\end{enumerate}

%-------------------------------------------------------------------------
\section{Related works}
\label{sec:related_works}

\paragraph{Text-to-image generation}

Text-to-image (T2I) generation models have evolved and iterated through several popular model architectures in the deep learning era. An early work is the Generative Adversarial Network (GAN)~\cite{goodfellow2014generative,brock2018biggan,karras2019stylegan}, which trains a generator for image generation and a discriminator to distinguish between real and generated images, in parallel (also see  \cite{zhang2017stackgan, xu2018attngan,zhu2019dm,qiao2019mirrorgan,li2019controllable,tao2022df} among others).
Another category of generation models develops from variational auto-encoders (VAEs)~\cite{kingma2013auto,higgins2016betavae,van2017vqvae}, which optimize evidence lower bound (ELBO) for the likelihood of the image data. 
% For T2I generation, StackGAN~\cite{zhang2017stackgan} uses a two-stage GAN to first generate low-resolution images based on text embedding and then refine the low-resolution images to a higher resolution. AttnGAN~\cite{xu2018attngan} utilizes an attentional model so that the image generation can focus on some keywords. DM-GAN~\cite{zhu2019dm} uses a dynamic memory module to refine any fuzzy image contents during generation. MirrowGAN~\cite{qiao2019mirrorgan} proposes a text-to-image-to-text training framework with a global-local collaborative attentive module. ControlGAN~\cite{li2019controllable} proposes a word-level spatial and channel-wise attention-driven generator that allows for fine-grained control of generated content using words. DF-GAN~\cite{tao2022df} fuses text information and visual feature maps through multiple text-image fusion layers. 
%

% 
Most recently, Diffusion Models (DMs)~\cite{sohl2015diffusion,ho2020denoisingdiffusion,nichol2021glide,rombach2022stablediffusion} have emerged as the state-of-the-art (SOTA) for Image Generation~\cite{dhariwal2021diffusionbeatsgans}. DMs are trained to generate images progressively from random noise, with the ability to capture more diversity than GANs and achieve good sample quality~\cite{dhariwal2021diffusionbeatsgans}. Latent Diffusion Models \cite{rombach2022stablediffusion} are a further refinement that performs the diffusion process in a compact latent space for more efficiency. 

% \TODO{Deepak or Gang or Feng: can you help write a little more for diffusion models?}

\vspace{-4mm}
\paragraph{Text-to-image evaluation and reward models}

There has been much recent work on evaluation of text-to-image models along many dimensions~\cite{xu2023imagereward,kirstain2023pickapic,huang2023t2ibench, wu2023hps,wu2023humanhpsv2,otani2023toward,lee2023holistic,Cho2023VPT2I}. \citet{xu2023imagereward} collect a human preference dataset by requesting users to rank multiple images and rate them according to their quality. They trained a reward model ImageReward for human preference learning, and proposed Reward Feedback Learning (ReFL) for tuning diffusion models with the ImageReward model. \citet{kirstain2023pickapic} built a web application to collect human preferences by asking users to choose the better image from a pair of generated images, resulting in a dataset called Pick-a-Pic with more than 500K examples generated by T2I models such as Stable Diffusion 2.1, Dreamlike Photoreal 2.05, and Stable Diffusion XL variants. They leveraged the human preference dataset to train a CLIP-based~\cite{radford2021clip} scoring function, called PickScore, to predict human preferences. \citet{huang2023t2ibench} proposed a benchmark called T2I-CompBench for evaluating text-to-image models, which consists of 6,000 text prompts describing attribute binding, object relationships, and complex compositions. They utilized multiple pretrained vision-language models such as CLIP~\cite{radford2021clip} and BLIP~\cite{li2022blip} to calculate multiple evaluation metrics. \citet{wu2023hps,wu2023humanhpsv2} collected a large scale dataset of human choices on generated images and utilized the dataset to train a classifier that outputs a Human Preference Score (HPS). They showed improvement in image generation by tuning Stable Diffusion with the HPS. 
Recently, Lee~\cite{lee2023holistic} proposed a holistic evaluation for T2I models with multiple fine-grained metrics. 

Despite these valuable contributions, most existing works only use binary human ratings or preference ranking for construction of feedback/rewards, and lack the ability to provide detailed actionable feedback such as implausible regions of the image, misaligned regions, or misaligned keywords on the generated images.  One recent paper related to our work is \citet{zhang2023perceptual}, which collected a dataset of artifact regions for image synthesis tasks, trained a segmentation-based model to predict artifact regions, and proposed a region inpainting method for those regions. However, the focus of their work is artifact region only, while in this paper, we collected rich feedback for T2I generation containing not only artifact regions, but also misalignment regions, misaligned keywords, and four fine-grained scores from multiple aspects. To the best of our knowledge, this is the first work on heterogeneous rich human feedback for text-to-image models.

% \subsection{Learning from human feedback}

% It has been shown that language models (LMs) can be significantly improved to align with human intent using reinforcement learning from human feedback (RLHF)~\cite{ouyang2022InstructGPT}. It works by first finetuning an LM with labeler-written prompts and responses, and then collecting a dataset of rankings of model outputs, training a reward model with the ranking data, and using reinforcement learning (RL) to optimize the LM against the reward model. %Moreover, some works provide more fine-grained feedback to the LLMs by providing token-level feedback or even rewriting the generated text, which significantly boosts the LLM performance. 
% Moreover, it is shown that fine-grained human feedback provides better rewards for LMs training~\cite{wu2023fine-grained}. 

% Recent works on evaluating or improving text-to-image models have drawn inspiration from the RLHF for LLMs. For example, Xu~\etal~\cite{xu2023imagereward} collected human ratings on generated images and trained a reward model called ImageReward to improve Stable Diffusion. Fan~\etal~\cite{fan2023reinforcement} proposed a method called DPOK to finetune T2I diffusion models with human feedback using RL. 

%-------------------------------------------------------------------------

\section{Collecting rich human feedback}
\label{sec:data_collection}

\begin{figure}[t!]
    \centering
    \includegraphics[width=0.8\linewidth]{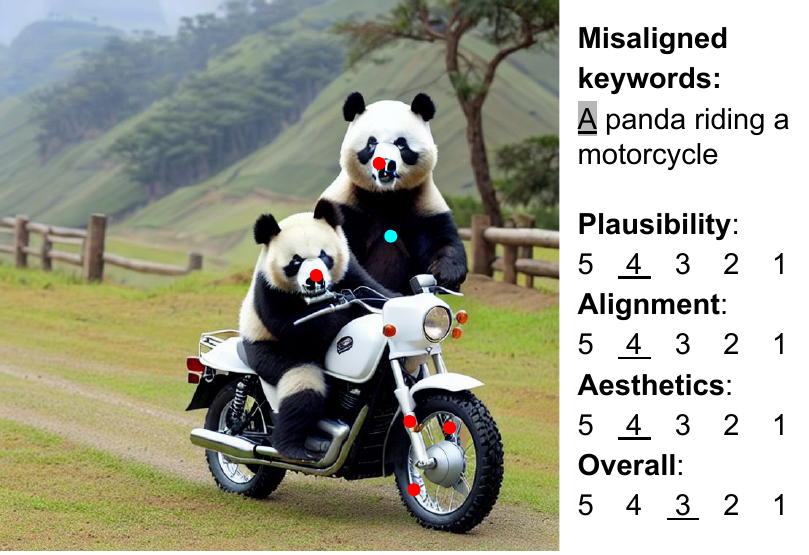}
    \vspace{-3mm}
    \caption{\textbf{An illustration of our annotation UI}. Annotators mark points on the image to indicate artifact/implausibility regions (red points) or misaligned regions (blue points) w.r.t the text prompt. Then, they click on the words to mark the misaligned keywords (underlined and shaded) and choose the scores for plausibility, text-image alignment, aesthetics, and overall quality (underlined).}
    \label{fig:annotation}
    \vspace{-3mm}
\end{figure}

\subsection{Data collection process}

In this section, we discuss our procedure to collect the RichHF-18K dataset, which includes two heatmaps (artifact/implausibility and misalignment), four fine-grained scores (plausibility, alignment, aesthetics, and overall score), and one text sequence (misaligned keywords).

% We consider the following annotations: 1) plausibility score, 2) image-text alignment score, 3) aesthetic score, 4) overall score, 5) image regions that have artifacts, implausible/unrealistic representations, 6) image regions that are misaligned with the text description, 7) keywords in the text description that are missing or misrepresented in the image.  

%Why did we choose these annotations?

For each generated image, the annotators are first asked to examine the image and read the text prompt used to generate it. 
% They are instructed to look up any unknown words in the text prompt using an internet search engine to make sure they understand the prompt. 
Then, they mark points on the image to indicate the location of any implausibility/artifact or misalignment w.r.t the text prompt. 
% Since implausibility/artifact regions have an area while points have no area, 
% To mark an area, annotators are instructed to imagine a disk-shape region centered at each point, where the disk has a fixed radius of around 1/20 of the image height. Annotators use points to mark the misalignment regions. They further mark the misaligned keywords in the text prompt by clicking on the keywords. 
The annotators are told that each marked point has an ``effective radius'' (1/20 of the image height), which forms an imaginary disk centering at the marked point. In this way, we can use a relatively small amount of points to cover the image regions with flaws. 
% The default state of each keyword is `aligned', indicating this keyword is properly represented in the generated image. A click on the keyword changes the state to `misaligned'. One more click changes the state to `unsure', meaning the keyword is not understood or the image is too blurry to determine if the keyword is properly represented. One more click changes the state back to `aligned', and so on. 
Lastly, annotators label the misaligned keywords and the four types of scores for plausibility, image-text alignment, aesthetic, and overall quality, respectively, on a 5-point Likert scale. Detailed definitions of image implausibility/artifact and misalignment can be found in the supplementary materials. 
We designed a web UI, as shown in \cref{fig:annotation}, to facilitate data collection. More details about data collection process can be found in the supplementary materials. 

\begin{figure*}[t!]
    \centering
    \includegraphics[width=\linewidth]{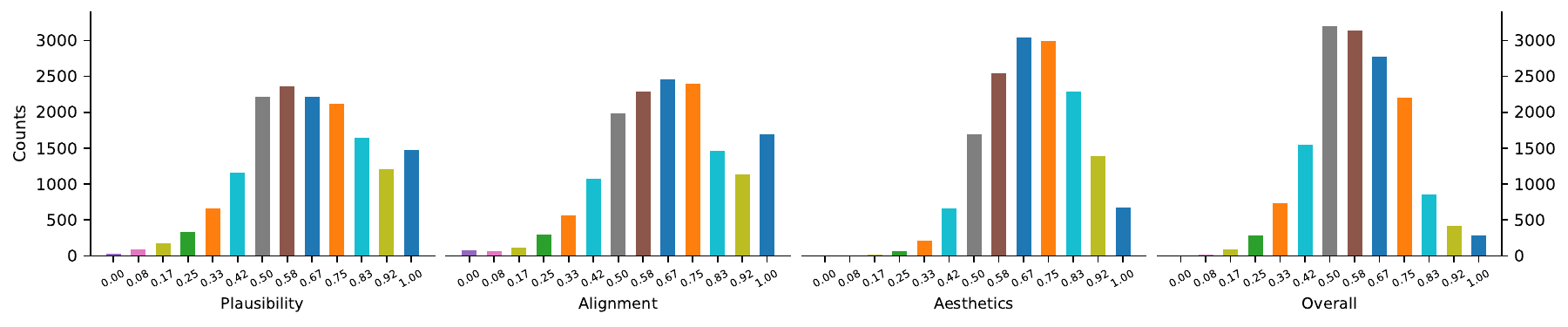}
    \vspace{-8mm}
    \caption{Histograms of the \textbf{average scores} of image-text pairs in the training set.}
    \label{fig:scores_hist}
    \vspace{-4mm}
\end{figure*}

\begin{figure*}[t!]
    \centering
    \includegraphics[width=0.8\linewidth]{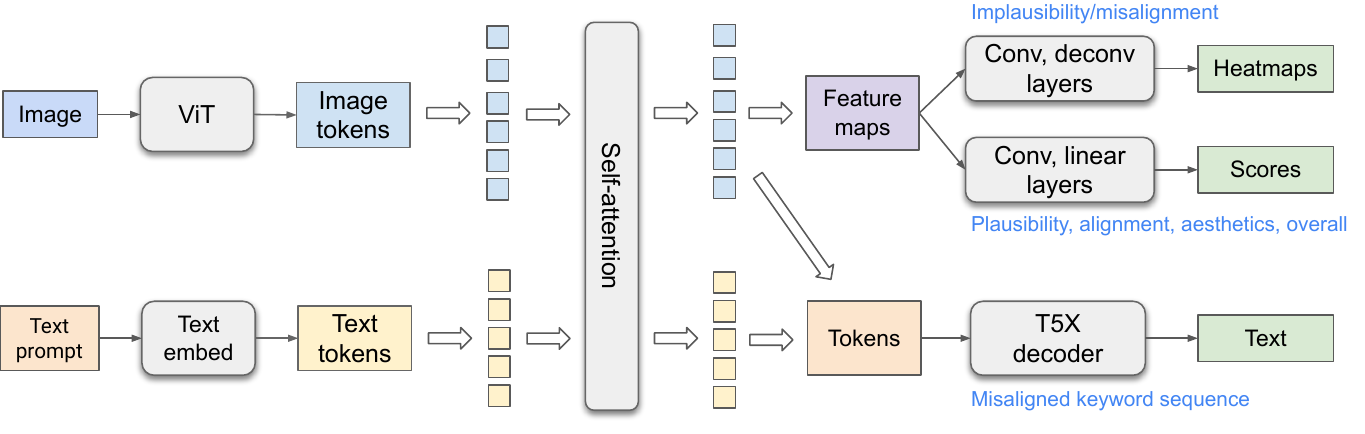}
    \vspace{-4mm}
    \caption{Architecture of our rich feedback model. Our model consists of two streams of computation: one vision and one text stream. We perform self-attention on the ViT-outputted image tokens and the Text-embed module-outputted text tokens to fuse the image and text information. The vision tokens are reshaped into feature maps and mapped to heatmaps and scores. The vision and text tokens are sent to a Transformer decoder to generate a text sequence.}
    \label{fig:model}
    \vspace{-5mm}
\end{figure*}

% We created detailed annotation guidelines
% %with more than 100 slides 
% to instruct the annotators regarding the annotation steps, interactions with the web UI, examples of different types of implausibility, artifacts, and misalignment. All the annotators (27 in total) are trained with the annotation guidelines and calibrated, before they perform the annotation in order to reduce annotation discrepancy and improve quality. 
% % Moreover, we randomly select samples for examination and provide feedback to the annotators on a daily basis.
% % \TODO{Youwei: definitions and subtlety in the image issues.}
% Our annotation took around 3,000 rater-hours in total. 
% More details about our data collection process are in the supplementary.
% % To improve the effectiveness of the collected dataset and control the time spent on annotation, we filter out any image-text pairs that have a text prompt with less than 3 words or more than 20 words. We also filter out non-English prompts or any prompts containing emoji. 
% % Prior to performing the final evaluation, we iterated upon the instructions for the annotators and calibrated their annotations using a pilot set of 100 image-text pairs, until satisfying annotation quality is achieved.

\subsection{Human feedback consolidation}

To improve the reliability of the collected human feedback on generated images, each image-text pair is annotated by three annotators. We therefore need to consolidate the multiple annotations for each sample. For the scores, we simply average the scores from the multiple annotators for an image to obtain the final score. For the misaligned keyword annotations, we perform majority voting to get the final sequence of indicators of aligned/misaligned, using the most frequent label for the keywords. For the point annotations, we first convert them to heatmaps for each annotation, where each point is converted to a disk region (as discussed in the last subsection) on the heatmap, and then we compute the average heatmap across annotators. 
% Specifically, for the points annotated by a annotator on an image, we create a disk of radius 1/20 of the image height for each point. The pixels on the disks have a value of 1 and the pixels outside the disks have a value of 0 (the maximum pixel value is 1 so the pixels in the overlapping part of the disks still get a value of 1), resulting in a heatmap for each annotator for an image. We average the heatmaps for all annotators for an image, producing the final heatmap. 
The regions with clear implausibility are likely to be annotated by all
annotators and have a high value on the final average heatmap.

\subsection{RichHF-18K: a dataset of rich human feedback}

We select a subset of image-text pairs from the Pick-a-Pic dataset for data annotation. Although our method is general and applicable to any generated images, we choose the majority of our dataset to be photo-realistic images, due to its importance and wider applications. Moreover, we also want to have balanced categories across the images. 
% have clearer definitions for plausibility/artifact for the image content. For example, cartoon or concept art photos can contain complex characters that are hard to determine whether they are visual flaws/artifacts or not. 
To ensure balance, we utilized the PaLI visual question answering (VQA) model~\cite{pali} to extract some basic features from the Pick-a-Pic data samples. Specifically, we asked the following questions for each image-text pair in Pick-a-Pic. 1) Is the image photorealistic? 2) Which category best describes the image? Choose one in `human', `animal', `object', `indoor scene', `outdoor scene'. PaLI's answers to these two questions are generally reliable under our manual inspection. 
% By visual examinations of the answers by the PaLI VQA model, we found that for questions 1) and 2), the answers are mostly correct. 
We used the answers to sample a diverse subset from Pick-a-Pic, resulting in 17K image-text pairs. We randomly split the 17K samples into two subsets, a training set with 16K samples and a validation set with 1K samples. The distribution of the attributes of the 16K training samples is shown in the supplementary materials. Additionally, we collect rich human feedback on the unique prompts and their corresponding images from the Pick-a-Pic test set as our test set. In total, we collected rich human feedback on the 18K image-text pairs from Pick-a-Pic. Our RichHF-18K dataset consists of 16K training, 1K validation, and 1K test samples.

%(Show the distribution of the same for the whole Pick-a-Pic dataset?)

% To reduce the randomness in data annotation, each image-text pair was annotated by 3 annotators. Moreover, the annotators are instructed to skip any images that are inappropriate (\eg, containing nudity) and we remove the inappropriate images from the collected data by checking if they are skipped. 

\begin{figure}[t!]
    \centering
    \includegraphics[width=0.9\linewidth]{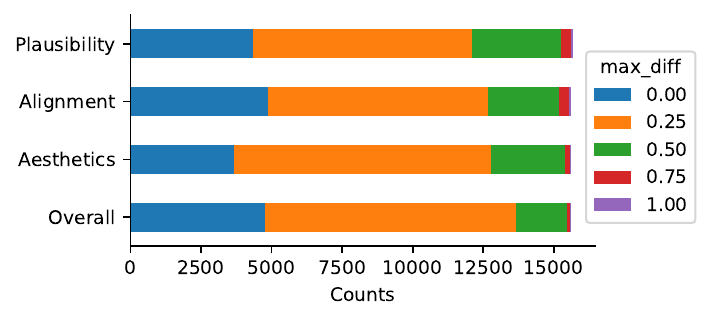}
    \vspace{-4mm}
    \caption{Counts of the samples with \textbf{maximum differences} of the scores in the training set.}
    \label{fig:max_diff}
    \vspace{-4mm}
\end{figure}

\vspace{-5px}
\subsection{Data statistics of RichHF-18K}

In this section, we summarize the statistics of the scores and perform the annotator agreement analysis for the scores. We standardize the scores $s$ with the formula $(s - s_{\text{min}}) / (s_{\text{max}} - s_{\text{min}})$ ($s_{\text{max}}=5$ and $s_{\text{min}}=1$) so that the scores lie in the range [0, 1]. 

The histogram plots of the scores are shown in~\cref{fig:scores_hist}. The distribution of the scores is similar to a Gaussian distribution, while the plausibility and text-image alignment scores have a slightly higher percentage of score 1.0. The distribution of the collected scores ensures that we have a reasonable number of negative and positive samples for training a good reward model. 

To analyze the rating agreement among annotators for an image-text pair, we calculate the maximum difference among the scores: $\text{max}_{\text{diff}} = \max(\text{scores}) - \min(\text{scores})$ where scores are the three score labels for an image-text pair. We plot the histogram of $\text{max}_{\text{diff}}$ in \cref{fig:max_diff}. We can see that around 25\% of the samples have perfect annotator agreement and around 85\% of the samples have good annotator agreement ($\text{max}_{\text{diff}}$ is less than or equal to 0.25 after the standardization or 1 in the 5-point Likert scale). 

% \TODO{Jordi: what is the annotator agreement of previous point-wise annotation? Can we claim our annotation of rich feedback together help improve the rating qaulity and annotator agreement?}

% \TODO{Compute the annotator agreement for heatmaps and misalignment sequence too. Probably do not have time now. Maybe add them to appendix.}

%Table~? shows the training attributes of each training datasets.

%-------------------------------------------------------------------------

\section{Predicting rich human feedback}
\subsection{Models}
\label{sec:models}

%\TODO{Need to polish this section.}

% \TODO{Youwei, Junfeng and Gang: Need to expand this section. Have subsections for model architecture, loss separately.}

% \TODO{Youwei, Junfeng and Gang: Mention two variants of our model: 1) Multi heads for heatmap (or score) 2) single head for heatmap  (or score), with heatmap/score type in the augmented prompt, which actually works better \eg, for artifact heatmap prediction.}

\subsubsection{Architecture}

The architecture of our model is shown in~\cref{fig:model}. We adopt a vision-language model based on ViT~\cite{vit} and T5X~\cite{t5x} models, inspired by the Spotlight model architecture~\cite{li2023spotlight}, but modifying both the model and pretraining datasets to better suit our tasks. We use a self-attention module~\cite{vaswani2017attention} among the concatenated image tokens and text tokens, similar to PaLI~\cite{pali}, as our tasks require bidirectional information propagation. The text information is propagated to image tokens for text misalignment score and heatmap prediction, while the vision information is propagated to text tokens for better vision-aware text encoding to decode the text misalignment sequence. To pretrain the model on more diverse images, we add the natural image captioning task on the WebLI dataset~\cite{pali} to the pretraining task mixture. 

 Specifically, the ViT takes the generated image as input and outputs image tokens as high-level representations. The text prompt tokens are embedded into dense vectors. The image tokens and embedded text tokens are concatenated and encoded by the Transformer self-attention encoder in T5X.
On top of the encoded fused text and image tokens, we use three kinds of predictors to predict different outputs. For heatmap prediction, the image tokens are reshaped into a feature map and sent through convolution layers, deconvolution layers, and sigmoid activation, and outputs implausibility and misalignment heatmaps. For score prediction, the feature map is sent through convolution layers, linear layers, and sigmoid activation, resulting in scalars as fine-grained scores. 

To predict the keyword misalignment sequence, the original prompt used to generate the image is used as text input to the model. A modified prompt is used as the prediction target for the T5X decoder. The modified prompt has a special suffix (`$\_0$') for each misaligned token, \eg, \textit{a yellow\_0 cat} if the generated image contains a black cat and the word \textit{yellow} is misaligned with the image. During evaluation, we can extract the misaligned keywords using the special suffix.

%we have two streams of computation: 1) the vision steam outputs heatmaps and scores for generated images, representing artifact/misaligned regions and fine-grained scores, respectively, and 2) the language stream decodes text tokens for misaligned keyword sequence. 

%In the vision stream, the ViT provides image tokens as a query (Q) and the T5X provides a key (K) and a value (V) from the text tokens. A multi-layer cross attention module performs cross attention on the Q, K, and V, infusing the text information into the vision tokens, where the output is reshaped to a feature map. 

\vspace{-0.5em}
\subsubsection{Model variants}

% As we need to predict four different scores, two different heatmaps, and the text misalignment sequence, we have explored two model variants: a multi-head model and a model with augmented prompts.

%We have many attributes to predict (four different scores, two different heatmaps, and the text misalignment sequence); as such, there are a variety of model structures we could consider to capture this diversity of signals. We focus on two model variants: a multi-headed model and a model with augmented prompts.

We explore two model variants for the prediction heads of the heatmaps and scores.

\vspace{-4.5mm}
\paragraph{Multi-head} A straightforward way to predict multiple heatmaps and scores is to use multiple prediction heads, with one head for each score and heatmap type. This will require seven prediction heads in total.

\vspace{-4.5mm}
\paragraph{Augmented prompt} Another approach is to use a single head for each prediction type, \ie, three heads in total, for heatmap, score, and misalignment sequence, respectively. To inform the model of the fine-grained heatmap or score type, we augment the prompt with the output type. More specifically, we prepend a task string (\eg, `implausibility heatmap') to the prompt for each particular task of one example and use the corresponding label as the training target. During inference, by augmenting the prompt with the corresponding task string, the single heatmap (score) head can predict different heatmaps (scores). As we show in the experiments, this augmented prompt approach can create task-specific vision feature maps and text encodings, which performs significantly better in some of the tasks.

\subsubsection{Model optimization}
We train the model with a pixel-wise mean squared error (MSE) loss for the heatmap prediction, and MSE loss for the score prediction.
For misalignment sequence prediction, the model is trained with teacher-forcing cross-entropy loss. The final loss function is the weighted combination of the heatmap MSE loss, score MSE loss, and the sequence teacher-forcing cross-entropy loss. 

%\TODO{Gang: Add Loss formula}

%#The whole model went through a pretraining stage where it was trained to output xxx?

%-------------------------------------------------------------------------

\subsection{Experiments}
\label{sec:experiments}

\begin{figure}[t!]
 \centering
  % This one contains human faces
  % \begin{subfigure}{0.24\textwidth}
  %   \includegraphics[width=\textwidth]{fig/heatmaps/artifact_map_e1a66d43-13dc-4e85-8b0b-c167eefef668.png}
  %   \caption{Image}
  % \end{subfigure}%
  % \hspace{2px}
  % \begin{subfigure}{0.24\textwidth}
  %   \includegraphics[width=\textwidth]{fig/heatmaps/artifact_map_e1a66d43-13dc-4e85-8b0b-c167eefef668-gt.png}
  %   \caption{Annotators}
  % \end{subfigure}%
  % \hspace{2px}
  % \begin{subfigure}{0.24\textwidth}
  %   \includegraphics[width=\textwidth]{fig/heatmaps/artifact_map_e1a66d43-13dc-4e85-8b0b-c167eefef668-maxogen.png}
  %   \caption{Our model}
  % \end{subfigure}%
  % \hspace{2px}
  % \begin{subfigure}{0.24\textwidth}
  %   \includegraphics[width=\textwidth]{fig/heatmaps/artifact_map_e1a66d43-13dc-4e85-8b0b-c167eefef668-resnet.png}
  %   \caption{ResNet baseline}
  % \end{subfigure}%
  
  % This is the backup
  % \\
   \begin{subfigure}{0.12\textwidth}
     \includegraphics[width=\textwidth]{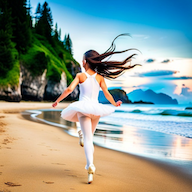}
    \caption{Image}
  \end{subfigure}%
  %\hspace{2px}
  \begin{subfigure}{0.12\textwidth}
    \includegraphics[width=\textwidth]{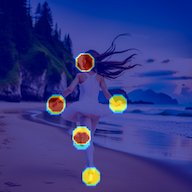}
    \caption{GT}
  \end{subfigure}%
  %\hspace{2px}
  \begin{subfigure}{0.12\textwidth}
    \includegraphics[width=\textwidth]{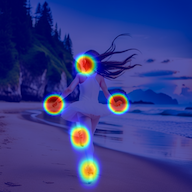}
    \caption{Our model}
  \end{subfigure}%
  %\hspace{2px}
  \begin{subfigure}{0.12\textwidth}
    \includegraphics[width=\textwidth]{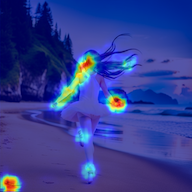}
    \caption{ResNet-50}
  \end{subfigure}%

\vspace{-3mm}
\caption{Examples of implausibility heatmaps. Prompt: \textit{photo of a slim asian little girl ballerina with long hair wearing white tights running on a beach from behind nikon D5}}
\label{fig:fidelity-map}
\vspace{-2mm}
\end{figure}

\begin{figure}[t!]
 \centering
  % This one contains human faces
  % \begin{subfigure}{0.24\textwidth}
  %   \includegraphics[width=\textwidth]{fig/heatmaps/misalignment_map_16a98806-aa12-493b-ab48-23721fa22be6.png}
  %   \caption{Image}
  % \end{subfigure}%
  % \hspace{2px}
  % \begin{subfigure}{0.24\textwidth}
  %   \includegraphics[width=\textwidth]{fig/heatmaps/misalignment_map_16a98806-aa12-493b-ab48-23721fa22be6-gt.png}
  %   \caption{Annotators}
  % \end{subfigure}%
  % \hspace{2px}
  % \begin{subfigure}{0.24\textwidth}
  %   \includegraphics[width=\textwidth]{fig/heatmaps/misalignment_map_16a98806-aa12-493b-ab48-23721fa22be6-maxogen.png}
  %   \caption{Our model}
  % \end{subfigure}%
  % \hspace{2px}
  % \begin{subfigure}{0.24\textwidth}
  %   \includegraphics[width=\textwidth]{fig/heatmaps/misalignment_map_16a98806-aa12-493b-ab48-23721fa22be6-clip.png}
  %   \caption{CLIP baseline}
  % \end{subfigure}%
  % This is the backup
  %\\
  \begin{subfigure}{0.12\textwidth}
    \includegraphics[width=\textwidth]{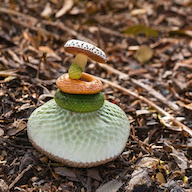}
    \caption{Image}
  \end{subfigure}%
  %\hspace{2px}
  \begin{subfigure}{0.12\textwidth}
    \includegraphics[width=\textwidth]{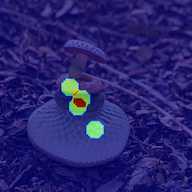}
    \caption{GT}
  \end{subfigure}%
  %\hspace{2px}
  \begin{subfigure}{0.12\textwidth}
    \includegraphics[width=\textwidth]{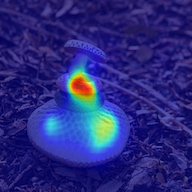}
    \caption{Our model}
  \end{subfigure}%
  %\hspace{2px}
  \begin{subfigure}{0.12\textwidth}
    \includegraphics[width=\textwidth]{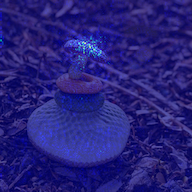}
    \caption{CLIP gradient}
  \end{subfigure}%

\vspace{-3mm}
\caption{Examples of misalignment heatmaps. Prompt:  \textit{A snake on a mushroom}.}
\label{fig:misalignment-map}
\vspace{-5mm}

% Photo of a cat eating a burger like a person
% An abandoned Segway in the forest
% inflatable pie floating down a river
% A little blonde girl busy eating a cup of yogurt on the counter top
% A Red Pigeon Sat on a Branch Reflecting on Existence
% A dog with a sign that reads Hello 
\end{figure}

\begin{table*}[t]
\centering
\resizebox{0.9\linewidth}{!}{%
\begin{tabular}{l|cc|cc|cc|cc}
\toprule
 & \multicolumn{2}{c|}{Plausibility}  & \multicolumn{2}{c|}{Aesthetics} & \multicolumn{2}{c|}{Text-image Alignment} & \multicolumn{2}{c}{Overall} \\
\midrule
& PLCC $\uparrow$ & SRCC $\uparrow$ & PLCC $\uparrow$ & SRCC $\uparrow$ & PLCC $\uparrow$ & SRCC $\uparrow$ & PLCC $\uparrow$ & SRCC $\uparrow$ \\
\midrule
ResNet-50 & 0.495 & 0.487 & 0.370 & 0.363 & 0.108 & 0.119 & 0.337 & 0.308 \\
PickScore (off-the-shelf) & 0.0098 & 0.0280 & 0.131 & 0.140 & 0.346 & 0.340 & 0.202 & 0.226 \\
% PickScore (fine-tuned) & 0.49795 & 0.48667 & 0.48194 & 0.47791 & \textbf{0.51635} & \textbf{0.52691} & 0.51821 & 0.51231 \\

CLIP (off-the-shelf)  & $-$ & $-$ & $-$ & $-$ & 0.185 & 0.130 & $-$ & $-$ \\
CLIP (fine-tuned) & 0.390 & 0.378 & 0.357 & 0.360 & 0.398 & 0.390 & 0.353 & 0.352 \\
Our Model (multi-head) &  0.666 & 0.654 & \textbf{0.605} & \textbf{0.591} & \textbf{0.487} & \textbf{0.500} & \textbf{0.582} & 0.561 \\
Our Model (augmented prompt) & \textbf{0.693} & \textbf{0.681} & 0.600 & 0.589 & 0.474 & 0.496 & 0.580 & \textbf{0.562} \\
% ResNet-50 & 0.49502 & 0.48713 & 0.37021 & 0.36308 & 0.10836 & 0.11935 & 0.33723 & 0.30788 \\
% CLIP &  &  &  &  &  0.18491 & 0.12996  &  \\
% Our Model (multi-head) &  0.66609 & 0.65392 & 0.60512 & 0.59132 & 0.48718 & 0.49956 & 0.58233 & 0.56052 \\
% Our Model (augmented prompt) & 0.69282 & 0.68106 & 0.59973 & 0.58945 & 0.47363 & 0.49643 & 0.58046 & 0.56207 \\
\bottomrule
\end{tabular}%
}
\vspace{-2mm}
\caption{Score prediction results on the test set.}
\label{tab:score_result}
\end{table*}

\begin{table*}[t]
\vspace{-2mm}
\centering
\resizebox{0.75\linewidth}{!}{%
\begin{tabular}{l|c|c|ccccc}
\toprule
 & All data & $GT=0$ & \multicolumn{5}{c}{$GT>0$} \\
\midrule
& MSE $\downarrow$ & MSE $\downarrow$ & CC $\uparrow$ & KLD $\downarrow$ & SIM $\uparrow$ & NSS $\uparrow$ & AUC-Judd $\uparrow$ \\
\midrule
ResNet-50 & 0.00996 & \textbf{0.00093} & 0.506 & 1.669 & 0.338 & 2.924 & 0.909 \\
Ours (multi-head) & 0.01216 & 0.00141 & 0.425 & 1.971 & 0.302 & 2.330 & 0.877 \\
Ours (augmented prompt) & \textbf{0.00920} & 0.00095 & \textbf{0.556} & \textbf{1.652} & \textbf{0.409} & \textbf{3.085} & \textbf{0.913} \\
% ResNet-50 & 0.00996 & 0.00093 & 0.50553 & 1.66916 & 0.33786 & 2.92368 & 0.90928 \\
% Ours (multi-head) & 0.01216 & 0.00141 & 0.42547 & 1.97131 & 0.30218 & 2.32973 & 0.87708 \\
% Ours (augmented prompt) & 0.00920 & 0.00095 & 0.55590 & 1.65192 & 0.40939 & 3.08488 & 0.91342 \\
\bottomrule
\end{tabular}%
}
\vspace{-2mm}
\caption{Implausibility heatmap prediction results on the test set. $GT=0$ refers to empty implausibility heatmap, \ie, no artifacts/implausibility (69 out of 995 test samples are empty), for ground truth. $GT>0$ refers to heatmaps with artifacts/implausibility, for ground truth.}
\label{tab:artifact_heatmap_result}
\end{table*}

\begin{table*}[t]
\centering
\resizebox{0.75\linewidth}{!}{%
\begin{tabular}{l|c|c|ccccc}
\toprule
 & All data & $GT=0$ & \multicolumn{5}{c}{$GT>0$} \\
\midrule
& MSE $\downarrow$ & MSE $\downarrow$ & CC $\uparrow$ & KLD $\downarrow$ & SIM $\uparrow$ & NSS $\uparrow$ & AUC-Judd $\uparrow$ \\
\midrule
CLIP gradient &  0.00817 & 0.00551 & 0.015 & 3.844 & 0.041 & 0.143 & 0.643 \\
Our Model (multi-head) & \textbf{0.00303} & 0.00015 & 0.206 & \textbf{2.932} & 0.093 & 1.335 & 0.838 \\
Our Model (augmented prompt) & 0.00304 & \textbf{0.00006} & \textbf{0.212} & 2.933 & \textbf{0.106} & \textbf{1.411} & \textbf{0.841} \\
% CLIP &  0.00817 & 0.00551 & 0.01543 & 3.84423 & 0.04134 & 0.14315 & 0.64329 \\
% Our Model (multi-head) & 0.00303 & 0.00015 & 0.20598 & 2.93238 & 0.09331 & 1.33500 & 0.83780 \\
% Our Model (augmented prompt) & 0.00304 & 0.00006 & 0.21152 & 2.93316 & 0.10632 & 1.41083 & 0.84120 \\
\bottomrule
\end{tabular}%
}
\caption{Text misalignment heatmap prediction results on the test set. $GT=0$ refers to empty misalignment heatmap, \ie, no misalignment (144 out of 995 test samples are empty), for ground truth. $GT>0$ refers to heatmaps with misalignment, for ground truth.}
\label{tab:misalignment_heatmap_result}
\end{table*}

\begin{table}[t]
\vspace{-2mm}
\centering
\resizebox{0.8\linewidth}{!}{%
\begin{tabular}{l|ccc}
\toprule
 & Precision & Recall & F1 Score  \\
\midrule
Multi-head & \textbf{62.9} & 33.0 & 43.3 \\
Augmented prompt & 61.3 & \textbf{34.1} & \textbf{43.9} \\
\bottomrule
\end{tabular}%
}
\vspace{-2mm}
\caption{Text misalignment prediction results on the test set.}
\label{tab:misalignment_seq_result}
% \vspace{-1mm}
\end{table}

\begin{figure*}[t!]
%\vspace{-2mm}
 \centering
  % \begin{subfigure}{0.24\textwidth}
  %   % This one contains human faces
  %   \includegraphics[align=t,width=0.95\textwidth]{fig/scores/72d5ee90-8e94-4608-aeca-4406e2de4aa6.png}
  %   \caption{GT: 0.333, Our model: 0.328 \\ Plausibility score. \\ }
  % \end{subfigure}%
  % \hspace{2px}
  %\begin{subfigure}{0.24\textwidth}
   % This one is a backup if the one above doesn't pass review
  %  \includegraphics[align=t,width=.95\textwidth]{fig/scores/615bb629-7460-48ce-b14c-87ee1d6ed5d7.png}
  %  %\caption{Prompt: \textit{All the letters of the greek alphabet}. \\ \\
  %  \caption{Prompt: \textit{Computer science students fighting with computer keyboards}.\\   
  %  Plausibility score. \\GT: 0.25, Our model: 0.236 \\
  %  Overall score. \\GT: 0.5, Our model: 0.341 }
  %  %\caption{}
  % \end{subfigure}%
  
  \begin{subfigure}{0.24\textwidth}
   % This one is a backup if the one above doesn't pass review
   \includegraphics[align=t,width=\textwidth]{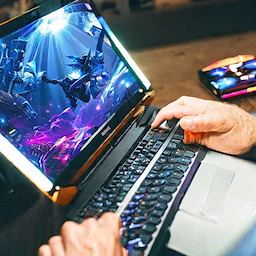}
   \caption{Prompt: \textit{gamer playing league of legends at night}.\\    \\
   Plausibility score. \\GT: 0.333, Our model: 0.410 \\
   Overall score. \\GT: 0.417, Our model: 0.457 }
   %\caption{}
  \end{subfigure}%
  % \begin{subfigure}{0.24\textwidth}
   % This one is a backup if the one above doesn't pass review
   % \includegraphics[align=t,width=.95\textwidth]{fig/scores/4904d2a1-41c9-4476-bbd4-b6e529ae135f.png}
   % \caption{Prompt: \textit{meditation under a rainbow during a thunderstorm}.\\ \\
   % Plausibility score. \\GT: 0.5, Our model: 0.448 \\
   % Overall score. \\GT: 0.583, Our model: 0.505 }
   %\caption{}
  % \end{subfigure}%
  \hspace{2px}
  \begin{subfigure}{0.24\textwidth}
    \includegraphics[align=t,width=\textwidth]{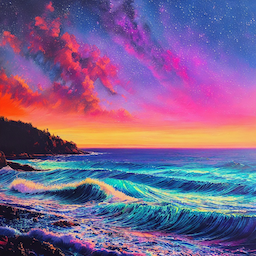}
    \caption{Prompt: \textit{An endless wavy ocean under a colorful night sky artistic painting pastel}.\\
    Plausibility score. \\GT: 1.0, Our model: 0.979 \\
    Overall score. \\GT 1.0, Our model: 0.848}
    %\caption{}
  \end{subfigure}%
  \hspace{2px}
  \begin{subfigure}{0.24\textwidth}
    % \includegraphics[align=t,width=.95\textwidth]{fig/scores/be3f702f-c418-4dee-9877-f964672d0c2d.png}
    % \caption*{Prompt: "heaven"\\ \\ \\
    % Text-image alignment score. \\GT: 0.25, Our model: 0.337 \\
    % Aesthetics score. \\ GT: 0.75, Our model: 0.713}
    \includegraphics[align=t,width=\textwidth]{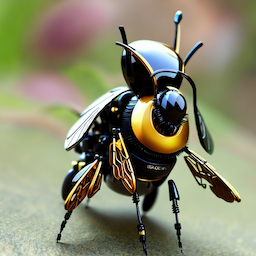}
    \caption{Prompt: \textit{Mechanical bee flying in nature electronics motors wires buttons lcd}.\\ 
    Text-image alignment score. \\GT: 0.583, Our model: 0.408 \\
    Aesthetics score. \\ GT: 0.75, Our model: 0.722}
    %\caption{}
  \end{subfigure}%
  \hspace{2px}
  % \begin{subfigure}{0.24\textwidth}
  %   \includegraphics[align=t,width=0.95\textwidth]{fig/scores/d1b93427-b2c1-4379-bf15-85bb3e561534.png}    
  %   \caption{Prompt: \textit{A needle-felted palm tree}.\\ \\
  %   Text-image alignment score. \\GT: 0.75, Our model: 0.988 \\
  %   Aesthetics score. \\ GT: 0.75, Our model: 0.961}
  %   %\caption{}
  % \end{subfigure}%
  \begin{subfigure}{0.24\textwidth}
    \includegraphics[align=t,width=\textwidth]{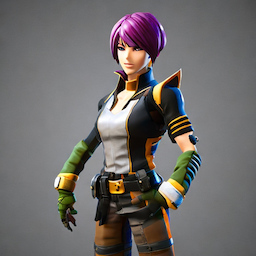}    
    \caption{Prompt: \textit{anime fortnite character}.\\ \\ \\
    Text-image alignment score. \\GT: 1.0, Our model: 0.897 \\
    Aesthetics score. \\ GT: 0.75, Our model: 0.713}
    %\caption{}
  \end{subfigure}%
  % \begin{subfigure}{0.24\textwidth}
  %   \includegraphics[align=t,width=0.95\textwidth]{fig/scores/1009f9e5-3d2c-4735-bad7-326064cb4d10.png}    
  %   \caption{Prompt: \textit{Renault Capture on a beach}.\\ \\ 
  %   Text-image alignment score. \\GT: 1.0, Our model: 
  %   0.877 \\
  %   Aesthetics score. \\ GT: 0.75, Our model: 0.720}
  %   %\caption{}
  % \end{subfigure}%

\caption{Examples of ratings. ``GT'' is the ground-truth score (average score from three annotators). }
\label{fig:scores}
\end{figure*}

\begin{figure*}[t!]
 \centering
  \begin{subfigure}{0.24\textwidth}
    \includegraphics[width=\textwidth]{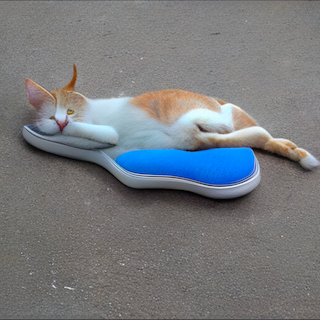}
    \caption {Muse~\cite{chang2023muse} before finetuning \label{fig:cat_before}}
  \end{subfigure}%
    \hspace{2px}
  \begin{subfigure}{0.24\textwidth}
    \includegraphics[width=\textwidth]{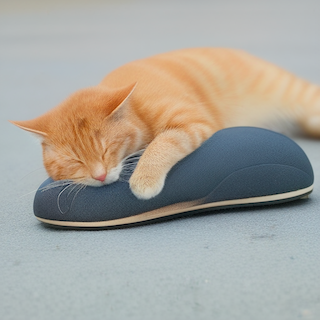}
    \caption{Muse~\cite{chang2023muse} after finetuning \label{fig:cat_after}}
  \end{subfigure}%
  \hspace{2px}
  \begin{subfigure}{0.24\textwidth}
    \includegraphics[width=\textwidth]{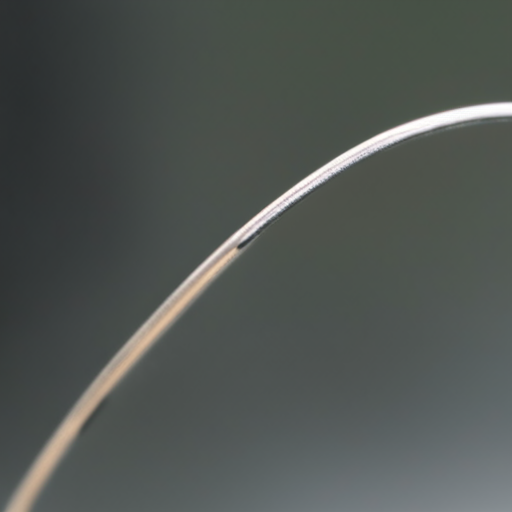}
    \caption{LD~\cite{rombach2022stablediffusion} without guidance \label{fig:coco_508248_before}}
  \end{subfigure}%
  \hspace{2px}
   \begin{subfigure}{0.24\textwidth}
    \includegraphics[width=\textwidth]{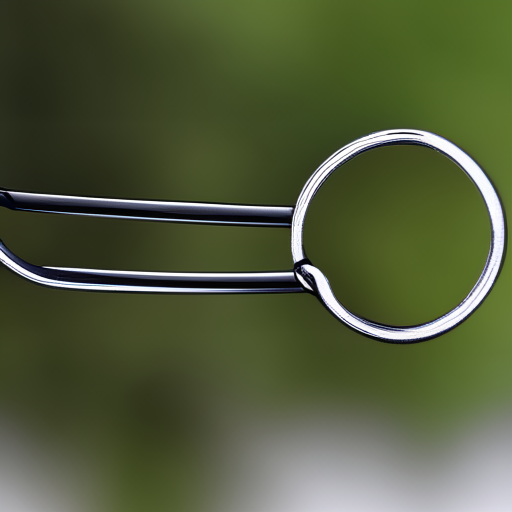}
    \caption{LD~\cite{rombach2022stablediffusion} after aesthetic guidance \label{fig:coco_508248_after}}
  \end{subfigure}%

  %  \begin{subfigure}{0.24\textwidth}
  %   \includegraphics[width=\textwidth]{fig/muse/coco_231290_before.png}
  %   \caption{Before finetuning \label{fig:coco_231290_before}}
  % \end{subfigure}%
  % \hspace{2px}

  % \begin{subfigure}{0.24\textwidth}
  %   \includegraphics[width=\textwidth]{fig/muse/coco_296324_before.png}
  %   \caption{Before finetuning \label{fig:coco_296324_before}}
  % \end{subfigure}%

  \caption{Examples illustrating the impact of RAHF on generative models.  (a-b): Muse~\cite{chang2023muse} generated images before and after finetuning with examples filtered by plausibility scores, prompt: \textit{A cat sleeping on the ground using a shoe as a pillow}. (c-d): Results without and with aesthetic score used as Classifier Guidance~\cite{bansal2023universal} on Latent Diffusion (LD)~ \cite{rombach2022stablediffusion}, prompt: \textit{a macro lens closeup of a paperclip.}}~\label{fig:muse_examples}
  \vspace{-4mm}
\end{figure*}

\begin{figure*}
\vspace{-2mm}
    \begin{subfigure}{\textwidth}
    \centering
        \includegraphics[width =0.8\textwidth]{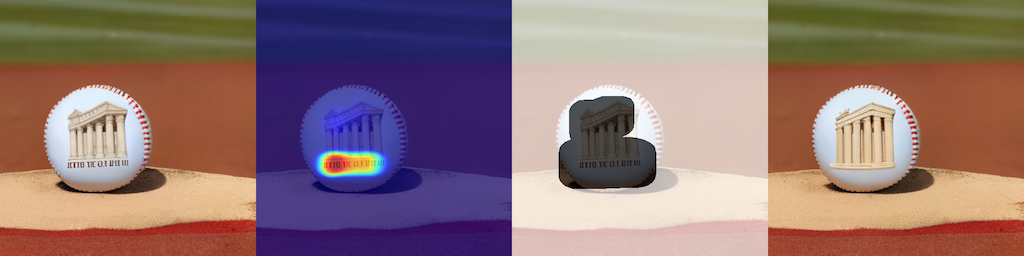}
        \caption{Prompt: \textit{a baseball with the parthenon on its cover, sitting on the pitcher's mound}}
    \end{subfigure} \\
    % \begin{subfigure}{\textwidth}
    %    \includegraphics[width = \textwidth]{fig/palm_caption_00000081_00_example.png}
    %    \caption{Prompt: \textit{A 3d printed sculpture of a cat made of iron and plastic, with arabic translation and ic gradients}.}
    % \end{subfigure} \\
    \begin{subfigure}{\textwidth}
    \centering
        \includegraphics[width = 0.8\textwidth]{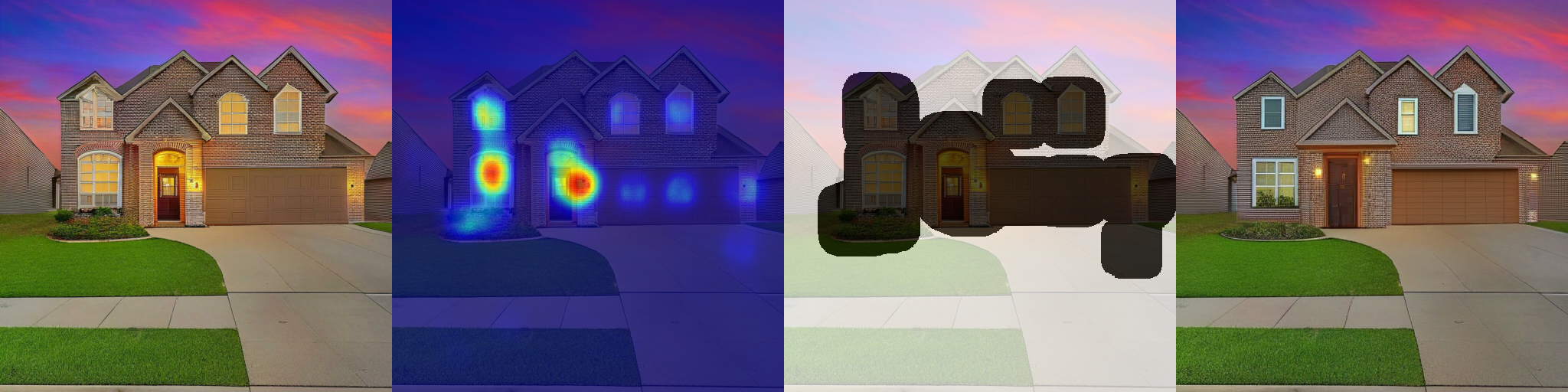} 
        \caption{Prompt: \textit{A photograph of a beautiful, modern house that is located in a quiet neighborhood. The house is made of brick and has a large front porch. It has a manicured lawn and a large backyard.}}
    \end{subfigure} \\
    % \begin{subfigure}{\textwidth}
    %     \includegraphics[width = \textwidth]{fig/palm_caption_00000042_05_example.png}
    %     \caption{Prompt: "a 1960s slide out camper with a blonde, white and red color scheme"}
    % \end{subfigure} \\
    %\begin{subfigure}{\textwidth}
    %    \includegraphics[width = \textwidth]{fig/palm_caption_00000433_05_example.png}
    %    \caption{Prompt: "a bathtub made of stainless steel, with a tabletop made of the same material, sitting in the middle of a field of flowers"}
    %\end{subfigure} \\
    %\begin{subfigure}{\textwidth}
    %    \includegraphics[width = \textwidth]{fig/palm_caption_00000275_01_example.png}
    %    \caption{Prompt: "a banana split with peanuts on top, sitting on a fruit stand in the airport lobby"}
    %\end{subfigure} \\

\vspace{-2em}
\caption{Region inpainting with Muse~\cite{chang2023muse} generative model. From left to right, the 4 figures are: original images with artifacts from Muse, predicted implausibility heatmaps from our model, masks by processing (thresholding, dilating) the heatmaps, and new images from Muse region inpainting with the mask, respectively.}
\label{fig:muse_inpating}
\end{figure*}

\begin{table}[]
\vspace{-2mm}
\centering
\resizebox{.9\columnwidth}{!}{
\begin{tabular}{@{}lccccc@{}}
\toprule
Preference & $\gg$ & $>$ & $\approx$ & $<$ & $\ll$\\ \midrule
Percentage & 21.5\% & 30.33\% & 31.33\% & 12.67\% & 4.17\% \\ \bottomrule
\end{tabular}
}
\vspace{-0.5em}
\caption{\textbf{Human Evaluation Results: Finetuned Muse vs original Muse model preference}: 
Percentage of examples where finetuned Muse is significantly better ($\gg$), slightly better ($>$), about the same ($\approx$), slightly worse ($<$), significantly worse ($\ll$) than original Muse.
Data was collected from 6 individuals in a randomized survey.}
\label{tab:humaneval}
\vspace{-2em}
\end{table}

\subsubsection{Experimental setup} 

% For the ViT of the reward model, we utilize the MaxViT, and for the text transformer, we exploit the T5X framework. Before training the model in our text-to-image reward modeling framework, we utilize some multi-modal vision-language tasks to pretrain the model. The pretraining tasks include image captioning and visual question answering. 
% Then, we finetune the model with our human feedback data using the three datasets including Pick-a-Pic, Mixed models, and ArtStation/Pexels, with a mixing rate of 1.0, 1.0, and 0.5, respectively. It means that in every 5 image-text pair inputs, there are 2 from Pick-a-Pic, 2 from Mixed models, and 1 from ArtStation/Pexels. 

Our model is trained on the 16K RichHF-18K training samples, and the hyperparameters were tuned using the model performance on the 1K RichHF-18K validation set. 
The hyperparameters setup can be found in supplementary material. %\cref{sec:hyperparameter}.

\vspace{-4.5mm}
\paragraph{Evaluation metrics}
For score prediction tasks, we report the Pearson linear correlation coefficient (PLCC) and Spearman rank correlation coefficient (SRCC), which are typical evaluation metrics for score predictions \cite{ke2021musiq}.
% We adopt SRCC instead of other correlation metrics since we are mostly interested in the ranking of the image-text pairs – whether one generated image is better than another one conditioned on the same text prompt.
For heatmap prediction tasks, a straightforward way to evaluate the results would be to borrow standard saliency heatmap evaluation metrics such as NSS/KLD~\cite{bylinskii2018different}. However, these metrics cannot be applied directly in our case as all these metrics assume the ground truth heatmap is not empty; yet in our case, empty ground truth is possible (\eg, for artifact/implausibility heatmap, it means the image does not have any artifact/implausibility). As such, we report MSE on all samples and on those with empty ground truth, respectively, and report saliency heatmap evaluation metrics like NSS/KLD/AUC-Judd/SIM/CC~\cite{bylinskii2018different} for the samples with non-empty ground truth.
% we report the MSE. We didn't use metrics like KLD/similarity as many ground truth heatmaps are empty, which causes the KLD/similarity inconsistent with human judgment. 
For the misaligned keyword sequence prediction, we adopt the token-level precision, recall, and F1-score. Specifically, the precision/recall/F1 scores are computed for the misaligned keywords over all the samples. %We use this metric because the token categories are imbalanced (put the statistics somewhere).

\vspace{-4.5mm}
\paragraph{Baselines}
For comparison, we finetune two ResNet-50 models~\cite{he2016resnet}, with multiple fully connected layers and deconvolution heads to predict the scores and heatmaps, respectively. We also use the off-the-shelf PickScore model~\cite{kirstain2023pickapic} to compute the PickScores and calculate the metrics w.r.t each of our four ground truth scores. We use the off-the-shelf CLIP model~\cite{radford2021clip} as a baseline to compute the cosine similarity of the image and text embeddings and use it to calculate the text-image alignment metric, as the CLIP cosine similarity is designed to reflect the alignment between images and prompts. Besides, we also fine-tune a CLIP model to predict the four types of scores using our training dataset. For misalignment heatmap prediction, we use CLIP gradient \cite{DBLP:journals/corr/SimonyanVZ13} map as a baseline. 

\vspace{-1em}
\subsubsection{Prediction result on RichHF-18K test set}

\paragraph{Quantitative analysis}
The experimental results of our model prediction on the four fine-grained scores, the implausibility heatmap, misalignment heatmap, and misalignment keyword sequence on our RichHF-18K test set are presented in \cref{tab:score_result}, \cref{tab:artifact_heatmap_result}, \cref{tab:misalignment_heatmap_result}, and \cref{tab:misalignment_seq_result} respectively. 

In both \cref{tab:score_result} and \cref{tab:misalignment_heatmap_result},  the two variants of our proposed model both significantly outperform ResNet-50 (or CLIP for text-image alignment score). Yet, in \cref{tab:artifact_heatmap_result}, the multi-head version of our model performs worse than ResNet-50,  but our augmented prompt version outperforms ResNet-50. The main reason might be that in the multi-head version, without augmenting the prediction task in the prompt, the same prompt is used for all the seven prediction tasks, and hence the feature maps and text tokens will be the same for all tasks. It might not be easy to find a good tradeoff among these tasks, and hence the performance of some tasks such as artifact/implausibility heatmap became worse. However, after augmenting the prediction task into a prompt, the feature map and text token can be adapted to each particular task with better results. Additionally, we note that misalignment heatmap prediction generally has worse results than artifact/implausibility heatmap prediction, possibly because misalignment regions are less well-defined, and the annotations may therefore be noisier.

\vspace{-4.5mm}
\paragraph{Qualitative examples}
We show some example predictions from our model for implausibility heatmap (\cref{fig:fidelity-map}), where our model identifies the regions with artifact/implausibility, and for misalignment heatmap (\cref{fig:misalignment-map}), where our model identifies the objects that don't correspond to the prompt.
\cref{fig:scores} shows some example images and their ground-truth and predicted scores.  More examples are in the supplementary material.
% Note that the score predictor is trained with SRCC loss besides L2 loss, and SRCC loss is mainly about the ranking instead of matching the exact value of the GT.

% Note that only relative scores were evaluated, and the results show that predicted scores don't span the whole $[0,1]$  range.

% The misalignment tasks (especially, heatmap prediction) are more difficult than others. Collecting the ground truth is hard, because misalignment can manifest in so many different ways: the object may be correct but in a wrong spot, or has a wrong color or another attribute, it could also be in a wrong relationship with other objects. Also, the annotators were not consistent in the way the objects are marked: whether it's fully covered in circles, or one dot per object. These factors contribute to the ground truth being noisy, and thus the model not to perform well.

% \paragraph{Generalization experiments}
% To test the generalization of our model, we hold out the images from one generative model in training and test the model performance on those images in evaluation. 

% \paragraph{Ablation on training datasets}
% In our initial experiments, we found that it is important to include real images and images from various generative models to improve the generalization of our reward model. To demonstrate this point, we train a reward model without the real images and report the results in xxx. We further remove images from the additional generative models for ablation and report the results in xxx.

% \section{Improve image generation with rich human feedback}
\section{Learning from rich human feedback}
% \TODO{Junfeng and Youwei: The model is trained on Pick-a-Pic (stable diffusion) data only, but LHF is done on Muse images. So this section is now both LHF exp and generalization exp. We will need to explain it clearly. }

In this section, we investigate whether the predicted rich human feedback (\eg, scores and heatmaps) can be used to improve image generation. To ensure that the gains from our RAHF model generalize across generative model families, we mainly use Muse~\cite{chang2023muse} as our target model to improve, which is based on a masked transformer architecture and thus different from the Stable Diffusion model variants
%used to produce the images 
in our RichHF-18K dataset. 
%As we demonstrate, our rich feedback prediction model can generalize quite well to images from unseen generative models.

\vspace{-3.5mm}
\paragraph{Finetuning generative models with predicted scores}

We first illustrate that finetuning with RAHF scores can improve Muse. First, we generate eight images for each of the \num{12564} prompts (the prompt set is created via PaLM 2~\cite{chowdhery2022palm,anil2023palm2} with some seed prompts) using the pre-trained Muse model. We predict RAHF scores for each image, and if the highest score for the images from each prompt is above a fixed threshold, it will be selected as part of our finetuning dataset. The Muse model is then finetuned with this dataset. This approach could be viewed as a simplified version of Direct Preference Optimization~\cite{fan2023reinforcement}. 

In~\cref{fig:muse_examples} (a)-(b), we show one example of finetuning Muse with our predicted plausibility score (threshold=0.8). 
%We show the image with the highest plausibility score for pre-trained and finetuned Muse respectively, illustrating  the benefits of using RAHF. 
To quantify the gain from Muse finetuning, we used 100 new prompts to generate images, and asked 6 annotators to perform side-by-side comparisons (for plausibility) between two images from the original Muse and the fine-tuned Muse respectively. The annotators choose from five possible responses (image A is significantly/slightly better than image B, about the same, image B is slightly/significantly better than image A), without knowledge of which model is used to generate the image A/B. The results in \cref{tab:humaneval} demonstrate that the finetuned Muse with RAHF plausibility scores possesses significantly fewer artifacts/implausibility than the original Muse. %as perceived by human annotators.

Moreover, in~\cref{fig:muse_examples}  (c)-(d), we show an example of using the RAHF aesthetic score as Classifier Guidance to the Latent Diffusion model~\cite{rombach2022stablediffusion}, similar to the approach in \citet{bansal2023universal}, demonstrating that each of the fine-grained scores can improve different aspects of the generative model/results.

% To show our reward model can be used to improve image generation, we finetune Muse with the scores predicted by our reward models. 

% \TODO{Gang and Jiao: Show examples of before vs after finetuning for the same prompts}.

%\TODO{Sarah and Jiao: Add human eval of finetuning results}.


% \usepackage{booktabs}

% % Please add the following required packages to your document preamble:
% % \usepackage{booktabs}
% \begin{table}[]
% \begin{tabular}{@{}l|l|l|l|l|l@{}}
% \toprule
%         & \textgreater{}\textgreater{} & \textgreater{} & =       & \textless{} & \textless{}\textless{} \\ \midrule
% Percent & 21.5\%                       & 30.33\%        & 31.33\% & 12.67\%     & 4.17\%                 \\ \bottomrule
% \end{tabular}
% \caption{Data was collected from 6 individuals in a randomized survey comparing images of before and after finetuning Muse with our reward models. 
% "\textgreater{}\textgreater{}" means finetuned with our models is significantly better than original, and "\textless{}\textless{}" means the pre-finetuned original image is significantly better.}
% \label{tab:humaneval}
% \end{table}

\vspace{-1em}
\paragraph{Region inpainting with predicted heatmaps and scores}

We demonstrate that our model's predicted heatmaps and scores can be used to perform region inpainting to improve the quality of generated images. For each image, we first predict implausibility heatmaps, then create a mask by processing the heatmap (using thresholding and dilating). Muse inpainting~\cite{chang2023muse} is applied within the masked region to generate new images that match the text prompt. Multiple images are generated, and the final image is chosen by the highest predicted plausibility score by our RAHF.

In~\cref{fig:muse_inpating}, we show several inpainting results with our predicted implausibility heatmaps and plausibility scores. As shown, more plausible images with fewer artifacts are generated after inpainting. Again, this shows that our RAHF generalizes well to images from a generative model very different from the ones whose images are used to train RAHF. More details and examples can be found in the supplementary material.
 
% \TODO{Nicholas: Add more examples of inpaiting}.

% \TODO{Sarah and Nicholas: Add human eval of inpainting results, after eval of finetuning results.}.

\section{Conclusions and limitations}

In this work, we contributed RichHF-18K, the first rich human feedback dataset for image generation. We designed and trained a multimodal Transformer to predict the rich human feedback, and demonstrated some instances to improve image generation with our rich human feedback. 

While some of our results are quite exciting and promising, there are several limitations to our work. First, the model performance on the misalignment heatmap is worse than that on the implausibility heatmaps, possibly due to the noise in the misalignment heatmap. It is somewhat ambiguous how to label some misalignment cases such as absent objects on the image. Improving the misalignment label quality is one of the future directions.
% maybe by changing how the misalignment is annotated now. 
Second, 
% even though models trained on RichHF-18K showed great generalization to unseen generative models like Muse, 
it would be helpful to collect more data on generative models beyond Pick-a-Pic (Stable Diffusion) and investigate their effect on the RAHF models. 
% \eg, to quantitatively investigate the effect of models trained across/among multiple sets (generative models). 
Moreover, while we present three promising ways to leverage our model to improve T2I generation, there is a myriad of other ways to utilize rich human feedback that can be explored, \eg, how to use the predicted heatmaps or scores as a reward signal to finetune generative models with reinforcement learning, and how to use the predicted heatmaps as a weighting map, or how to use the predicted misaligned sequences in learning from human feedback to help improve image generation, etc. 
We hope RichHF-18K and our initial models inspire quests to investigate these research directions in future work.

% to leverage diverse human feedback, \eg, bette

% besides the presented two ways of learning from rich human feedback, many different other ways to leverage rich human feedback can be explored, \eg, how to use the predicted heatmap as a weighting map, or how to use the predicted misaligned sequence to help improve image generation? All these will be our future work.

{
    \small
    \bibliographystyle{ieeenat_fullname}
    \bibliography{main}
}

% Per CVPR recommendation, the supplementary pages SHOULD be compiled with the main paper for cross reference.
\clearpage
\setcounter{page}{1}
\maketitlesupplementary

\section{Ethical conduct}
Our data collection has been approved by an Institutional Review Board. % This statement is sufficient per https://cvpr2023.thecvf.com/Conferences/2024/AuthorSuggestedPractices

\section{Data collection details}
\subsection{Image artifacts/implausibility definitions}
\label{sec:artifact_definitions}
\begin{enumerate}
    \item Distorted human/animal bodies/faces
    \begin{enumerate}
        \item Distorted/combined faces and/or body parts (unless specified in the text caption)
        \item Missing body parts (unless specified in the text caption)
        \item Additional body parts (unless specified in the text caption)
    \end{enumerate}
    \item Distorted objects (non human/animal)
    \begin{enumerate}
        \item Distorted objects (e.g., furniture, vehicles, buildings)  (unless specified in the text caption)
    \end{enumerate}
    \item Distorted/Nonsensical text
    \begin{enumerate}
        \item Text that is distorted, nonsensical, or misspelled  (unless specified in the text caption)
    \end{enumerate}
    \item Nonsensical Representations 
    \begin{enumerate}
        \item Representations that are unrealistic/nonsensical (unless specified in the text caption), or difficult to understand
    \end{enumerate}
    \item Excessive blurriness/lack of sharpness
    \begin{enumerate}
        \item The image contains excessive blurriness or quality that detracts from the overall image (focus on one part of the image is OK)
        \item The image contains a lack of definition/sharpness that detracts from the overall image
    \end{enumerate}
    \item Any other artifacts or implausibility not covered above
\end{enumerate}

\subsection{Text-image misalignment definitions and what-to-do}
Since we require the annotators to mark the misaligned words in the text prompt, we differentiate this part from \cref{sec:artifact_definitions} by including a what-to-do under each definition. 
\begin{enumerate}
    \item \textbf{Something is missing}: a human/animal/object specified in the text caption is missing in the image
    \begin{itemize}
        \item Click on that word of the human/animal/object in the text
    \end{itemize}
    \item \textbf{Incorrect attributes}: an attribute (e.g., color) of an object specified in the text is incorrect in the image
    \begin{itemize}
        \item Click on that word of the attribute in the text and click on the region of the object on the image
    \end{itemize}
    \item \textbf{Incorrect actions}: an action specified in the text caption is not represented in the image
    \begin{itemize}
        \item Click on that word of the action in the text and click on the region of the wrong actions on the image
    \end{itemize}
    \item \textbf{Incorrect numbers}: counts of humans/animals/objects in the image do not match those specified in the text
    \begin{itemize}
        \item Click on the number in the text
    \end{itemize}
    \item \textbf{Incorrect position}: the spatial position of two entities in the image does not match that specified in the text
    \begin{itemize}
        \item Click on the word of the position in the text
    \end{itemize}
    \item \textbf{Other}: any other inconsistency between text and image
    \begin{itemize}
        \item Click on the word of the inconsistency in the text
    \end{itemize}
\end{enumerate}

\subsection{Additional details}
\paragraph{Annotation guideline} To ensure the annotators understand the above definitions, we provide 4-10 examples for each definition of the annotation terms in the guideline. All of our annotators can read English and thus understand the text prompts. In some of the prompts, there are concepts or person names in the text prompts that are uncommon and may cause confusion to the annotators. Therefore, we instruct the annotators to do a quick search on the internet regarding any unfamiliar concepts in the text prompts and skip samples with confusing prompts full of strange concepts. 

\paragraph{Annotation interface} We designed a web UI to facilitate data collection with the following principles: 1) convenience for annotators to perform annotations, ideally within a short time for an image-text pair and, 2) allowing annotators to perform all annotations on the same UI, so that the fine-grained scores are based on the annotated regions and keywords. 
To this end, we created the interface as illustrated in \cref{fig:annotation}. The main UI consists of an image displayed on the left and a panel on the right, where the text prompt is shown at the top of the panel. Annotators are asked to first click on the image to annotate the artifact/implausible regions and misalignment regions, and then select the misaligned keywords and the fine-grained scores on the right of the panel. 

\paragraph{More details}
We created detailed annotation guidelines
%with more than 100 slides 
to instruct the annotators regarding the annotation steps, interactions with the web UI, examples of different types of implausibility, artifacts, and misalignment. All the annotators (27 in total) are trained with the annotation guidelines and calibrated, before they perform the annotation in order to reduce annotation discrepancy and improve quality. 
% Moreover, we randomly select samples for examination and provide feedback to the annotators on a daily basis.
% \TODO{Youwei: definitions and subtlety in the image issues.}
Our annotation took around 3,000 rater-hours in total. 
%More details about our data collection process are in the supplementary.
To improve the effectiveness of the collected dataset and control the time spent on annotation, we filter out any image-text pairs that have a text prompt with less than 3 words or more than 20 words. We also filter out non-English prompts or any prompts containing emoji. 
% Prior to performing the final evaluation, we iterated upon the instructions for the annotators and calibrated their annotations using a pilot set of 100 image-text pairs, until satisfying annotation quality is achieved.

\paragraph{Dataset size} 
Since the Pick-a-Pic v1 dataset contains some images and/or prompts that are inappropriate (\eg, containing nudity), we ask the annotators to mark these images with a special flag and skip the annotation. We filter out these inappropriate images and/or prompts during data post-processing. For this reason, the total number of images in our final training set is around 300 short of 16K.

\begin{figure}[t!]
    \centering
    \includegraphics[width=0.8\linewidth]{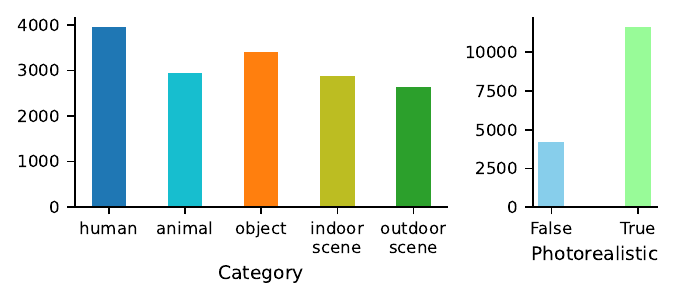}
    \vspace{-4mm}
    \caption{Histograms of the PaLI attributes of the images in the training set.}
    \label{fig:attributes}
    \vspace{-4mm}
\end{figure}

\paragraph{Additional details of data collection} The distribution of the attributes of the 16K training samples is shown in \cref{fig:attributes}. We can see a relatively balanced distribution of the types of content in the generated images in our dataset. %To analyze the rating agreement among annotators for an image-text pair, we calculate the maximum difference among the scores and plot the counts of the $\text{max}_{\text{diff}}$ for each score type in~\cref{fig:max_diff}.

\subsection{Discussions and limitations}

We choose \textbf{points over bounding boxes} in our region annotation because we find that points are much faster to mark and can have a reasonable coverage over image regions with various shapes when we specify an effective radius for each point as discussed in the main paper. 

As a limitation in our region/heatmap annotations, we notice there is an \textbf{over-annotation} issue in the artifacts/implausibility region annotation. Specifically, our annotators tend to annotate more human faces and hands on the images than necessary. One reason is that human faces and hands in the Pick-a-Pic dataset indeed have more artifacts/implausibility than other parts. Moreover, the annotators, as humans, may naturally pay more attention to human faces and hands, resulting in over-annotation of these parts. Nevertheless, the over-annotation issue is minor in our final dataset, as we strive to provide feedback to the annotators frequently to make them less nitpicking about human faces and hands. 

Another limitation is the \textbf{diversity} of the subjects in the prompts/images. The Pick-a-Pic dataset (and many others) is predominantly full of human, dog, and cat subjects. For this reason, it is challenging to find a very diverse dataset for annotation. We strive to make the subjects more diverse by using balanced categories as indicated by the PaLI attributes (as in \cref{fig:attributes}). We didn't choose more fine-grained categories for PaLI to test as there would be an endless list of subjects we could consider. Therefore, we leave the goal of annotating more diverse images/prompts in future works.

%\TODO{Examples for aggregating points as heatmaps}

\section{Experimental details}
\label{sec:hyperparameters}

\paragraph{Hyperparameters} The main model components consist of a ViT B16 encoder for image encoding, a T5 base encoder for mixing image and text tokens, and three predictors for score, heatmap, and text misalignment, respectively. The ViT B16 encoder uses a 16x16 patch size, 12 layers with 12 heads with a hidden dimension of 768, wherein the MLP has a hidden dimension of 3072. The T5 base encoder uses 12 layers with 12 heads with a hidden dimension of 768, wherein the MLP has a hidden dimension of 2048. The score predictor consists of four convolutional layers with layer norm and ReLU activation, and the filter size, kernel size, and strides are $[768, 384, 128, 64], [2, 2, 2, 2], [1, 1, 1, 1]$, respectively. Three dense layers of output sizes 2048, 1024, and 1, respectively, are used to generate a scalar with ReLU activation for the first two layers and sigmoid for the last. The heatmap predictor consists of two convolution layers with filter size, kernel size, and stride as $[768, 384], [3, 3], [1, 1]$, respectively. It then uses four de-convolution layers to up-sample to the required output size, with the filter size, kernel size, and stride as $[768, 384, 384, 192], [3, 3, 3, 3], [2, 2, 2, 2]$, respectively. Each de-convolution layer is with two read-out convolution layers of kernel size 3 and stride 1. Layer norm and ReLU are used for each layer. In the end, two read-out convolution layers and a final sigmoid activation are used to generate the heatmap prediction. The text predictor is implemented using a T5 base decoder with 12 layers of 12 heads, MLP dimension 2048, and hidden dimension 768. The output token length is 64.

We train the model on the datasets with a batch size of 256 for 20K iterations. We utilize the AdamW optimizer with a base learning rate of 0.015. We linearly increase the learning rate from 0 to the base learning rate in the first 2000 iterations, and then decrease the learning rate with a reciprocal square root scheduler w.r.t the number of iterations. We trained the model using 64 Google Cloud TPU v3 chips. 

\paragraph{Image augmentations} For each image, we randomly crop it by sampling a bounding box with 80\%-100\% width and 80\%-100\% height. The cropping is applied by 50\% chance and otherwise the original image is used. Note that we also crop the corresponding part of the implausibility heatmap and misalignment heatmap to match the cropped image. We then create an augmented version of the image by applying several random augmentations including random brightness (max delta 0.05), random contrast (random contrast factor between 0.8 and 1), random hue (max delta 0.025), random saturation (random saturation factor between 0.8 and 1) and random jpeg noise (jpeg quality between 70 and 100). By 10\% chance the augmented version is used instead of the original image. We convert the image to grayscale by 10\% probability as the final image.

%%%% Learning from rich human feedback
\paragraph{Finetuning generative models with predicted scores} 
To generate the training prompt set, we provide five hand-crafted seed prompts as examples and then ask PaLM 2~\cite{anil2023palm2} to generate similar textual prompts. We include additional instructions that specify the prompt length and the object category. We then explain why we do not use existing benchmark datasets for training. Theoretically,  we can get an infinite number of prompts using the prompt synthesis technique we proposed above. Existing datasets are 1) relatively small (e.g., TIFA~\cite{hu2023tifa} has 4k prompts, Davidsonian Scene Graph (DSG)~\cite{jaemin-cho-2023} has only 1k prompts), or 2) containing prompts that are simple and not diverse enough, for example, only measuring single objects in Parti benchmark~\cite{yu2022scaling}. This motivates us to synthesize a larger set of diverse prompts for training purposes. 
For the 100 prompts for our human evaluation, they are sampled from the existing benchmark: TIFA~\cite{hu2023tifa}. We only did our evaluation on 100 prompts due to the high cost of the human annotation.

\section{Additional qualitative examples}
\begin{figure*}[t!]
 \centering
  \begin{subfigure}{0.24\textwidth}
     \includegraphics[width=\textwidth]{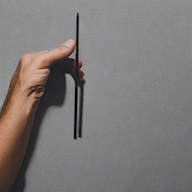}
     \includegraphics[width=\textwidth]{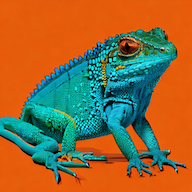}
     \includegraphics[width=\textwidth]{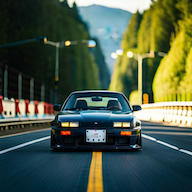}
     \includegraphics[width=\textwidth]{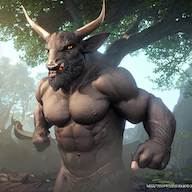}
     \includegraphics[width=\textwidth]{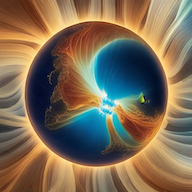}
     
    \caption{Image}
  \end{subfigure}%
  %\hspace{2px}
  \begin{subfigure}{0.24\textwidth}
    \includegraphics[width=\textwidth]{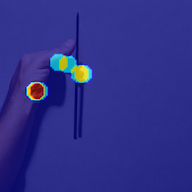}
    \includegraphics[width=\textwidth]{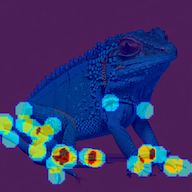}
     \includegraphics[width=\textwidth]{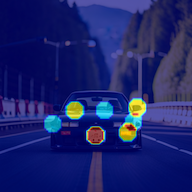}
     \includegraphics[width=\textwidth]{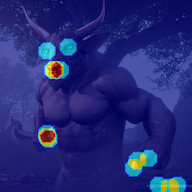}
     \includegraphics[width=\textwidth]{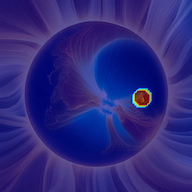}
    \caption{GT}
  \end{subfigure}%
  %\hspace{2px}
  \begin{subfigure}{0.24\textwidth}
    \includegraphics[width=\textwidth]{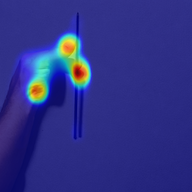}
    \includegraphics[width=\textwidth]{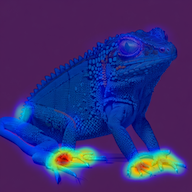}
     \includegraphics[width=\textwidth]{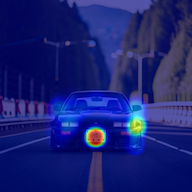}
     \includegraphics[width=\textwidth]{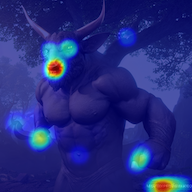}
     \includegraphics[width=\textwidth]{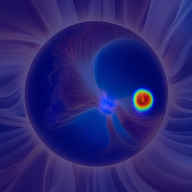}
    \caption{Our model}
  \end{subfigure}%
  %\hspace{2px}
  \begin{subfigure}{0.24\textwidth}
    \includegraphics[width=\textwidth]{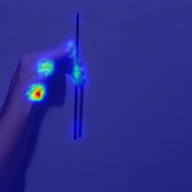}
    \includegraphics[width=\textwidth]{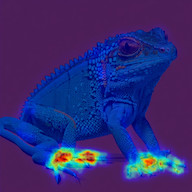}
     \includegraphics[width=\textwidth]{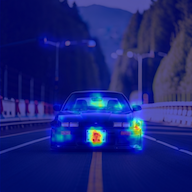}
     \includegraphics[width=\textwidth]{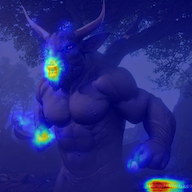}
     \includegraphics[width=\textwidth]{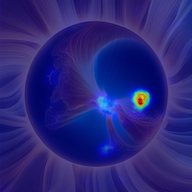}
    \caption{ResNet-50}
  \end{subfigure}%
\vspace{-3mm}
\caption{More examples of implausibility heatmaps}
\label{fig:fidelity-map_suppl_1}
\vspace{-2mm}
\end{figure*}

\cref{fig:fidelity-map_suppl_1} provides more examples of artifacts/implausibility heatmaps. We can see that our RAHF model can more accurately locate the positions of artifacts/implausibility on various subjects such as human hands, animals, vehicles, and concept arts. 

\begin{figure*}[t!]
 \centering
  \begin{subfigure}{\textwidth}
    \includegraphics[width=0.24\textwidth]{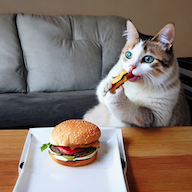}
    % \caption{Image}
    \includegraphics[width=0.24\textwidth]{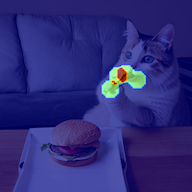}
    % \caption{GT}
    \includegraphics[width=0.24\textwidth]{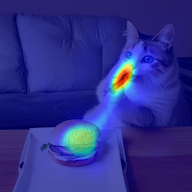}
    % \caption{Our model}
    \includegraphics[width=0.24\textwidth]{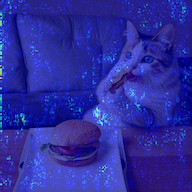}
    % \caption{CLIP gradient}
  \caption{Prompt: \textit{Photo of a cat eating a burger like a person}}
  \end{subfigure}%
  \\
  \begin{subfigure}{\textwidth}
    \includegraphics[width=0.24\textwidth]{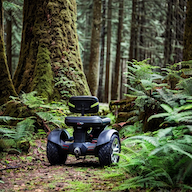}
    % \caption{Image}
    \includegraphics[width=0.24\textwidth]{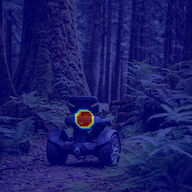}
    % \caption{GT}
    \includegraphics[width=0.24\textwidth]{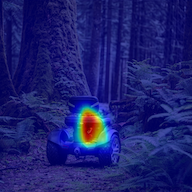}
    % \caption{Our model}
    \includegraphics[width=0.24\textwidth]{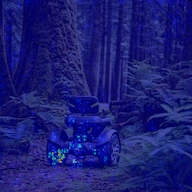}
    % \caption{CLIP gradient}
  \caption{Prompt: \textit{An abandoned Segway in the forest}}
  \end{subfigure}%
  \\
  \begin{subfigure}{\textwidth}
    \includegraphics[width=0.24\textwidth]{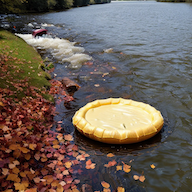}
    % \caption{Image}
    \includegraphics[width=0.24\textwidth]{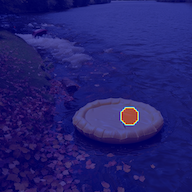}
    % \caption{GT}
    \includegraphics[width=0.24\textwidth]{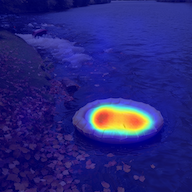}
    % \caption{Our model}
    \includegraphics[width=0.24\textwidth]{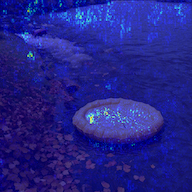}
    % \caption{CLIP gradient}
  \caption{Prompt: \textit{inflatable pie floating down a river}}
  \end{subfigure}%
  \\
  \begin{subfigure}{\textwidth}
    \includegraphics[width=0.24\textwidth]{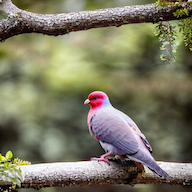}
    % \caption{Image}
    \includegraphics[width=0.24\textwidth]{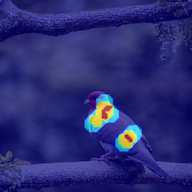}
    % \caption{GT}
    \includegraphics[width=0.24\textwidth]{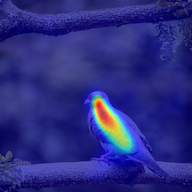}
    % \caption{Our model}
    \includegraphics[width=0.24\textwidth]{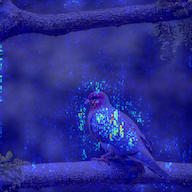}
    % \caption{CLIP gradient}
  \caption{Prompt: \textit{A Red Pigeon Sat on a Branch Reflecting on Existence}}
  \end{subfigure}%
  \\
  Image \hspace{3.5cm} GT \hspace{3cm}  Our model \hspace{2.5cm}  CLIP gradient

\caption{More examples of misalignment heatmaps.}
\label{fig:misalignment-map_suppl}

% A dog with a sign that reads Hello 
\end{figure*}

\cref{fig:misalignment-map_suppl} provides more examples of misalignment heatmaps. We can see that our RAHF model can more accurately locate the positions of misalignment on various subjects such as animals, objects, and different outdoor scenes. For example, our model can identify the subtle difference between the real handlebar of a Segway and the one depicted in the image.

\begin{figure*}[t!]
 \centering
  \begin{subfigure}{0.24\textwidth}
   \includegraphics[align=t,width=\textwidth]{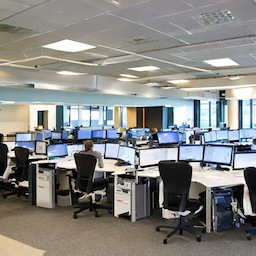}
   %\caption{Prompt: \textit{All the letters of the greek alphabet}. \\ \\
   \caption{Prompt: \textit{Computer science students fighting with computer keyboards}.\\   
   Plausibility score. \\GT: 0.25, Our model: 0.236 \\
   Overall score. \\GT: 0.5, Our model: 0.341 }
   %\caption{}
  \end{subfigure}%
  \hspace{2px}
  \begin{subfigure}{0.24\textwidth}
   \includegraphics[align=t,width=\textwidth]{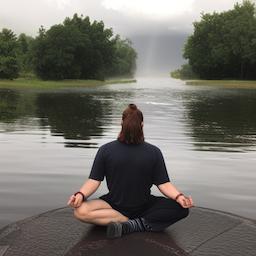}
   \caption{Prompt: \textit{meditation under a rainbow during a thunderstorm}.\\ \\
   Plausibility score. \\GT: 0.5, Our model: 0.448 \\
   Overall score. \\GT: 0.583, Our model: 0.505 }
  \end{subfigure}%
  \hspace{2px}
  \begin{subfigure}{0.24\textwidth}
    \includegraphics[align=t,width=\textwidth]{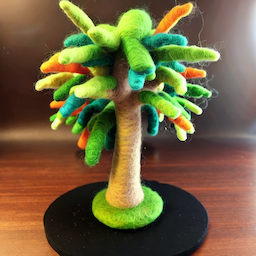}    
    \caption{Prompt: \textit{A needle-felted palm tree}.\\ \\
    Text-image alignment score. \\GT: 0.75, Our model: 0.988 \\
    Aesthetics score. \\ GT: 0.75, Our model: 0.961}
    %\caption{}
  \end{subfigure}%
  \hspace{2px}
  \begin{subfigure}{0.24\textwidth}
    \includegraphics[align=t,width=\textwidth]{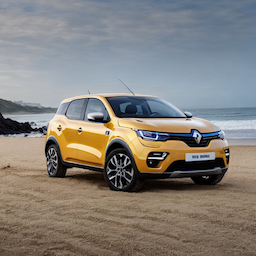}    
    \caption{Prompt: \textit{Renault Capture on a beach}.\\ \\ 
    Text-image alignment score. \\GT: 1.0, Our model: 
    0.877 \\
    Aesthetics score. \\ GT: 0.75, Our model: 0.720}
    %\caption{}
  \end{subfigure}%
  \\
  \begin{subfigure}{0.24\textwidth}
    \includegraphics[align=t,width=\textwidth]{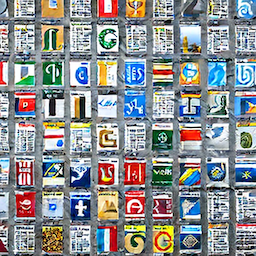}    
    \caption{Prompt: \textit{all the letters of the greek alphabet}.\\ 
    Plausibility score. \\GT: 0.167, Our model: 0.331 \\
    Overall score.  \\ GT: 0.250, Our model: 0.447 }
    %\caption{}
  \end{subfigure}%
  \hspace{2px}
  \begin{subfigure}{0.24\textwidth}
    \includegraphics[align=t,width=\textwidth]{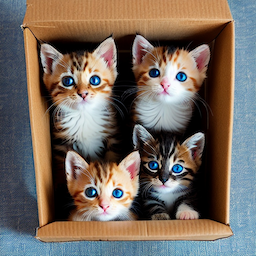}    
    \caption{Prompt: \textit{a kittens in box}.\\ \\ 
    Plausibility score. \\GT: 0.75, Our model: 0.851 \\
    Overall score.  \\ GT: 0.75, Our model: 0.855}
    %Text-image alignment score. \\GT: 0.917, Our model: 0.940 \\
    %Aesthetics score. \\ GT: 0.75, Our model: 0.877}
    %\caption{}
  \end{subfigure}%
  \hspace{2px}
  \begin{subfigure}{0.24\textwidth}
    \includegraphics[align=t,width=\textwidth]{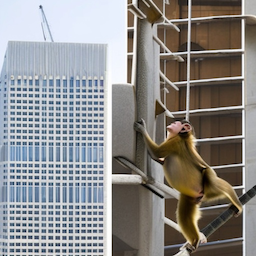}    
    \caption{Prompt: \textit{monkey climbing a skyscraper}.\\ 
    Text-image alignment score. \\GT: 0.833, Our model: 0.536 \\
    Aesthetics score. \\ GT: 0.583, Our model: 0.467}
    %\caption{}
  \end{subfigure}%
  \hspace{2px}
  \begin{subfigure}{0.24\textwidth}
    \includegraphics[align=t,width=\textwidth]{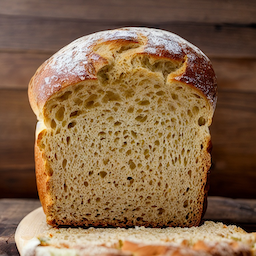}    
    \caption{Prompt: \textit{bread}.\\ \\ 
    Text-image alignment score. \\GT: 1.0, Our model: 0.975  \\
    Aesthetics score. \\ GT: 1.0, Our model: 0.984 }
    %\caption{}
  \end{subfigure}%

  % backup images
  % \\
  % \begin{subfigure}{0.24\textwidth}
  %   \includegraphics[align=t,width=0.95\textwidth]{fig/scores/a0b214e0-f6c0-4df1-9d56-34481692ffac.png}    
  %   \caption{Prompt: \textit{}.\\ \\ 
  %   Text-image alignment score. \\GT: , Our model:  \\
  %   Aesthetics score. \\ GT: , Our model: }
  %   %\caption{}
  % \end{subfigure}%
  % \begin{subfigure}{0.24\textwidth}
  %   \includegraphics[align=t,width=0.95\textwidth]{fig/scores/e383601b-5cc6-42bd-9044-f66cb8294c0e.png}    
  %   \caption{Prompt: \textit{}.\\ \\ 
  %   Text-image alignment score. \\GT: , Our model:  \\
  %   Aesthetics score. \\ GT: , Our model: }
  %   %\caption{}
  % \end{subfigure}%
  % \begin{subfigure}{0.24\textwidth}
  %   \includegraphics[align=t,width=0.95\textwidth]{fig/scores/f63adacc-5acd-40f9-b0b8-76b8916fc2c1.png}    
  %   \caption{Prompt: \textit{}.\\ \\ 
  %   Text-image alignment score. \\GT: , Our model:  \\
  %   Aesthetics score. \\ GT: , Our model: }
  %   %\caption{}
  % \end{subfigure}%

\caption{Examples of ratings. ``GT'' is the ground-truth score (average score from three annotators). }
\label{fig:scores_suppl}
\end{figure*}

\cref{fig:scores_suppl} provides more examples of score predictions, where our RAHF model predicts scores that are quite close to the ground truth score from human evaluation.

\begin{figure*}[t!]
    \begin{subfigure}{0.46\linewidth}
    \centering
    \includegraphics[width=0.8\linewidth]{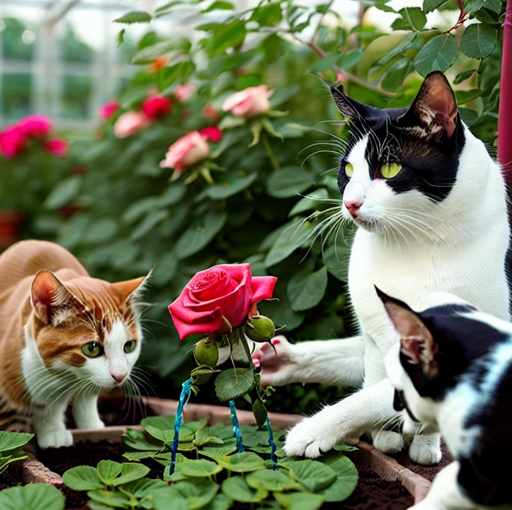}
    \caption{The prompt is: \textit{Two cats watering roses in a greenhouse}. The ground truth labels \textit{two, watering, greenhouse} as misaligned keywords and our model predicts \textit{two, greenhouse} as misaligned keywords. \phantom{This text will be invisible, just to occupy some spaces.}}
    \label{fig:text-example1}
    \end{subfigure}%
    \hspace{10px}
    \begin{subfigure}{0.46\linewidth}
    \centering
    \includegraphics[width=0.8\linewidth]{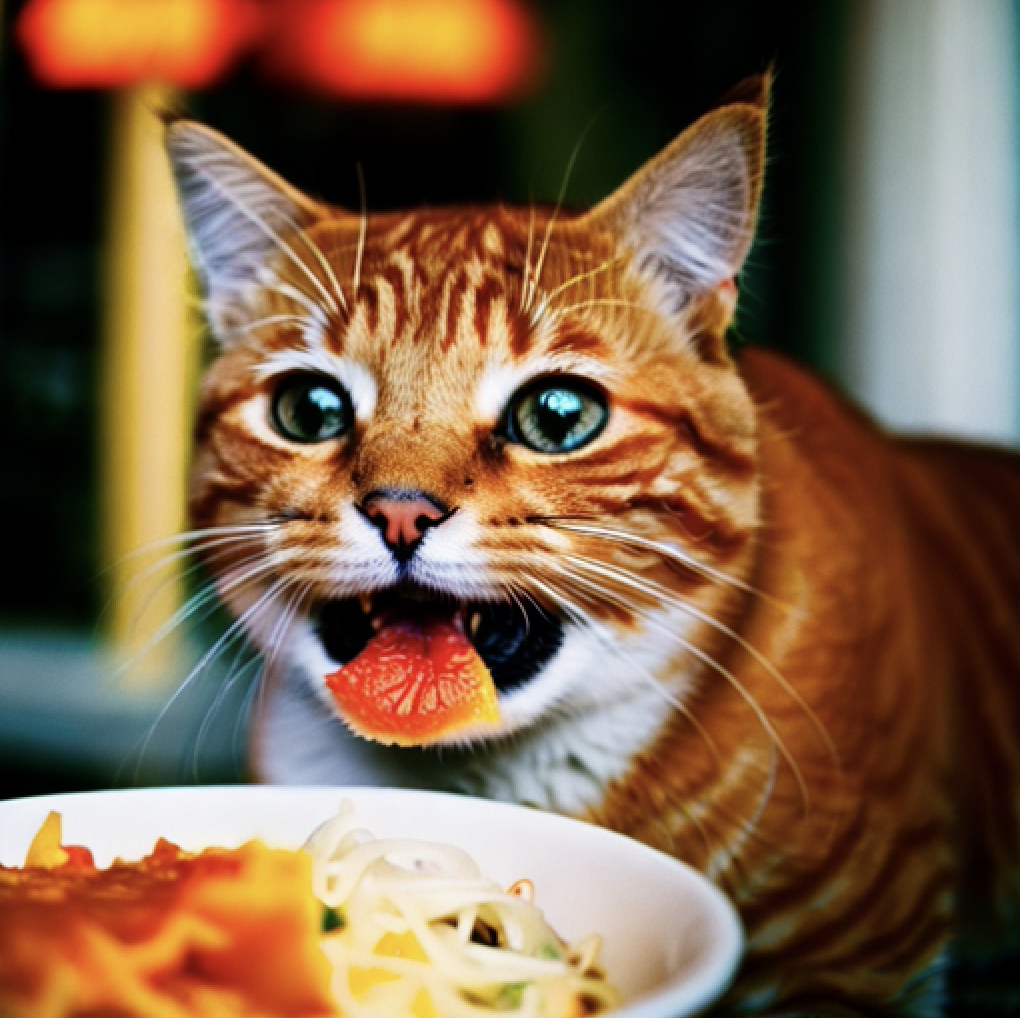}
    \caption{The prompt is: \textit{A close up photograph of a fat orange cat with lasagna in its mouth, shot on Leica M6}. The ground truth labels \textit{fat, lasagna, Leica, M6} as misaligned keywords and our model predicts \textit{lasagna, Leica, M6} as misaligned keywords.}
    \label{fig:text-example2}
    \end{subfigure}
    \caption{Examples for text misalignment prediction.}
    \label{fig:text-seq-prediction}
\end{figure*}

\cref{fig:text-seq-prediction} provides examples for the misaligned keywords prediction, which shows that our RAHF model can predict the majority of the misaligned keywords marked by human annotators.

\begin{figure*}[t!]
 \centering
  \begin{subfigure}{0.24\textwidth}
    \includegraphics[width=\textwidth]{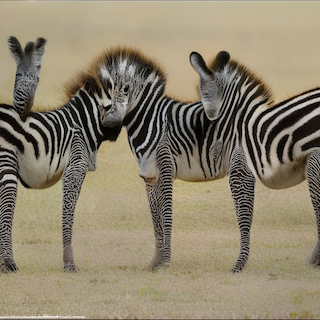}
    \caption {Muse before finetuning \label{fig:cat_before}}
  \end{subfigure}%
    \hspace{2px}
  \begin{subfigure}{0.24\textwidth}
    \includegraphics[width=\textwidth]{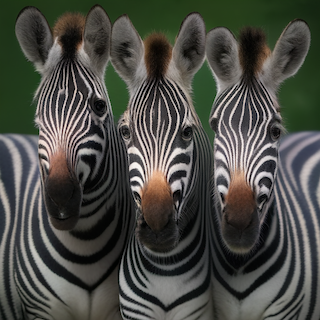}
    \caption{Muse after finetuning \label{fig:cat_after}}
  \end{subfigure}%
  \hspace{2px}
   \begin{subfigure}{0.24\textwidth}
    \includegraphics[width=\textwidth]{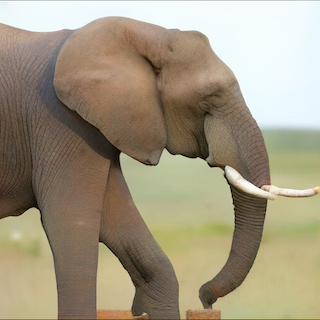}
    \caption{Muse before finetuning \label{fig:coco_184070_before}}
  \end{subfigure}%
  \hspace{2px}
   \begin{subfigure}{0.24\textwidth}
    \includegraphics[width=\textwidth]{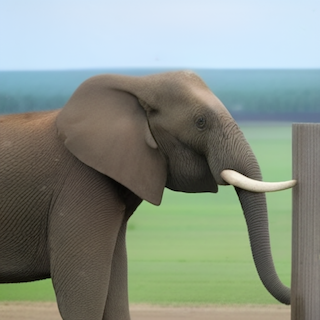}
    \caption{Muse after finetuning \label{fig:coco_184070_after}}
  \end{subfigure}%
  \hspace{2px}
  \begin{subfigure}{0.24\textwidth}
    \includegraphics[width=\textwidth]{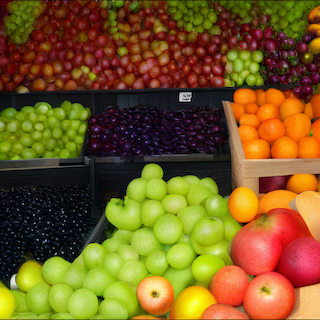}
    \caption{Muse before finetuning \label{fig:coco_296324_before}}
  \end{subfigure}%
    \hspace{2px}
  \begin{subfigure}{0.24\textwidth}
    \includegraphics[width=\textwidth]{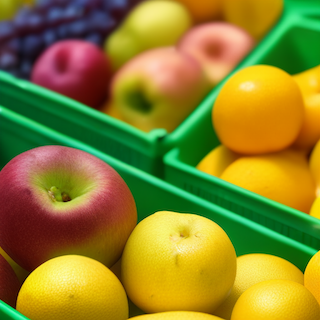}
    \caption{Muse after finetuning \label{fig:coco_296324_after}}
  \end{subfigure}%
  \hspace{2px}
  \begin{subfigure}{0.24\textwidth}
    \includegraphics[width=\textwidth]{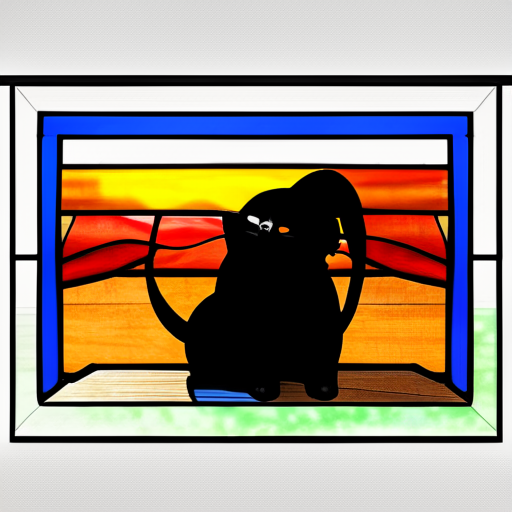}
    \caption{LD without guidance \label{fig:coco_508248_before}}
  \end{subfigure}%
  \hspace{2px}
   \begin{subfigure}{0.24\textwidth}
    \includegraphics[width=\textwidth]{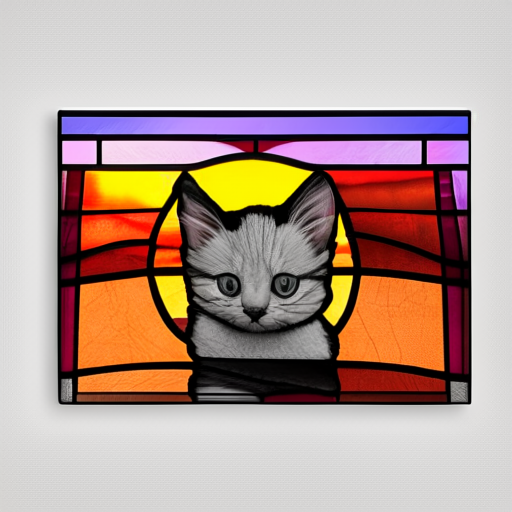}
    \caption{LD with overall guidance \label{fig:coco_508248_after}}
  \end{subfigure}%
  \caption{More examples illustrating the impact of RAHF on generative models.  (a-f): Muse~\cite{chang2023muse} generated images before and after finetuning with examples filtered by plausibility scores. Prompt: (a-b): \textit{Three zebras are standing together in a line.} (c-d): \textit{An elephant scratching it's neck on a post.} (e-f): \textit{Apples, lemons, grapes, oranges and other fruits in crates.} (g-h): Results without and with overall score used as Classifier Guidance~\cite{bansal2023universal} on Latent Diffusion (LD)~ \cite{rombach2022stablediffusion}, prompt: \textit{Kitten sushi stained glass window sunset fog.}}~\label{fig:muse_examples_suppl}
  %\vspace{-4mm}
\end{figure*}

\cref{fig:muse_examples_suppl} provides more examples of the comparison before and after finetuning Muse with examples selected based on the predicted scores by our RAHF model and examples of using RAHF model predicted overall score as Classifier Guidance. We can see enhanced image quality of the generation from the finetuned Muse model and the Latent Diffusion model, which highlights the potential of improving T2I generation with our reward model. 

\begin{figure*}
    % \begin{subfigure}{\textwidth}
    %     \includegraphics[width = \textwidth]{fig/palm_caption_00000398_04_example.png}
    %     \caption{Prompt: \textit{A baseball with the parthenon on its cover, sitting on the pitcher's mound.}}
    % \end{subfigure} \\
    \begin{subfigure}{\textwidth}
       \includegraphics[width = \textwidth]{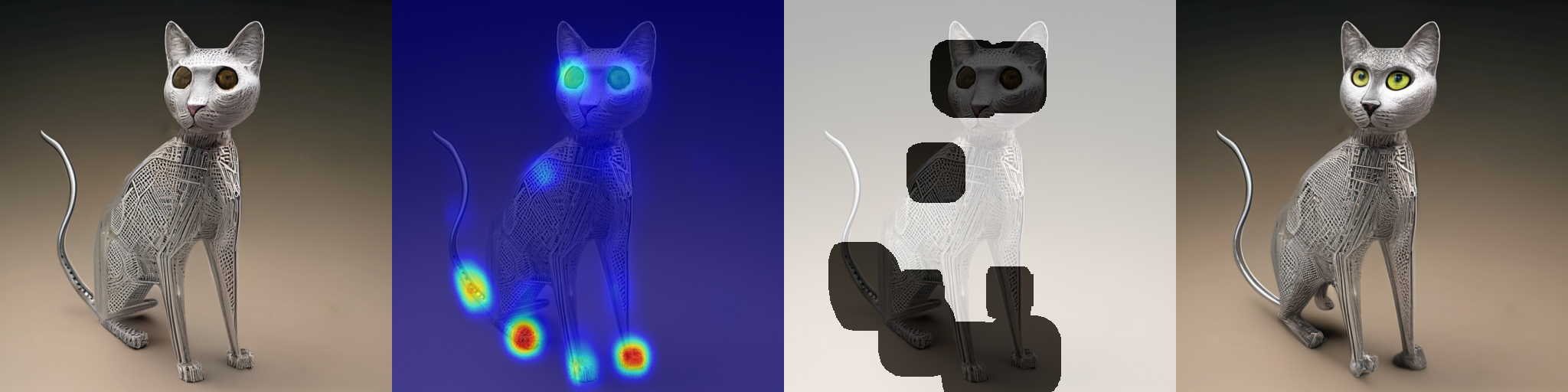}
       \caption{Prompt: \textit{A 3d printed sculpture of a cat made of iron and plastic, with arabic translation and ic gradients}.}
    \end{subfigure} \\
    % \begin{subfigure}{\textwidth}
    %     \includegraphics[width = \textwidth]{fig/ads_studio_000149_example.png} 
    %     \caption{Prompt: \textit{A photograph of a beautiful, modern house that is located in a quiet neighborhood. The house is made of brick and has a large front porch. It has a manicured lawn and a large backyard.}}
    % \end{subfigure} \\
    \begin{subfigure}{\textwidth}
        \includegraphics[width = \textwidth]{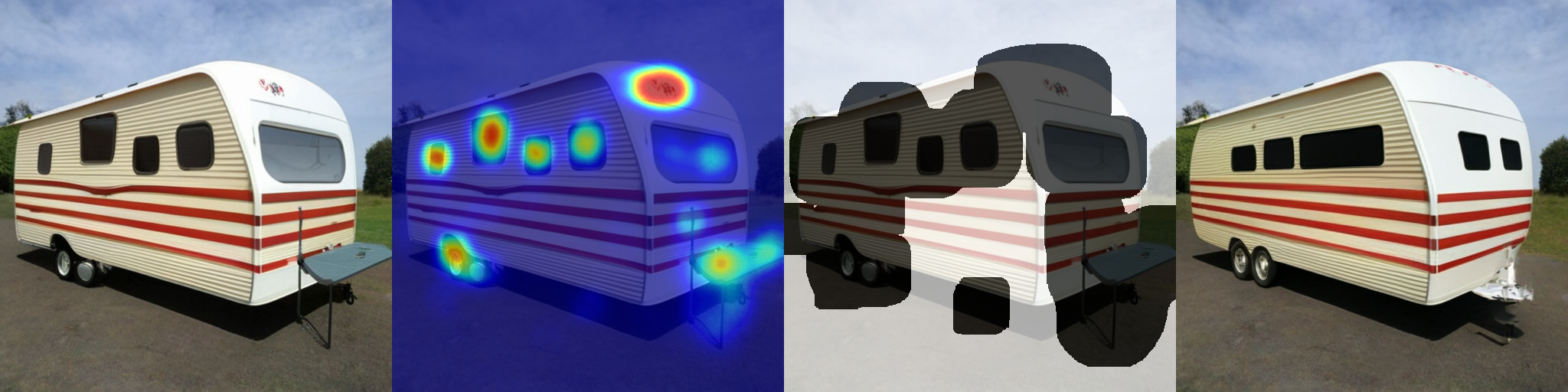}
        \caption{Prompt: \textit{A 1960s slide out camper with a blonde, white and red color scheme}}
    \end{subfigure} \\
    %\begin{subfigure}{\textwidth}
    %    \includegraphics[width = \textwidth]{fig/palm_caption_00000433_05_example.png}
    %    \caption{Prompt: "a bathtub made of stainless steel, with a tabletop made of the same material, sitting in the middle of a field of flowers"}
    %\end{subfigure} \\
    %\begin{subfigure}{\textwidth}
    %    \includegraphics[width = \textwidth]{fig/palm_caption_00000275_01_example.png}
    %    \caption{Prompt: "a banana split with peanuts on top, sitting on a fruit stand in the airport lobby"}
    %\end{subfigure} \\

\vspace{-2em}
\caption{Region inpainting with Muse~\cite{chang2023muse} generative model. From left to right, the 4 figures are: original images with artifacts from Muse, predicted implausibility heatmaps from our model, masks by processing (thresholding, dilating) the heatmaps, and new images from Muse region inpainting with the mask, respectively.}
\label{fig:muse_inpating_suppl}
\end{figure*}

\cref{fig:muse_inpating_suppl} provides more examples of Muse inpainting with the predicted masks (converted from heatmaps) by our RAHF model, where the inpainted regions are significantly improved in plausibility.

%\iffalse
%\fi

% Skip this part
\iffalse
\section{Human evaluation for LHF with Latent Diffusion model}

In Tab. \ref{tab:ld_humaneval}, we include human evaluation for comparing images generated by the original Latent Diffusion (LD) model \cite{rombach2022stablediffusion} and the LD model with RAHF overall score as Classifier Guidance \cite{bansal2023universal}. We can see the LD model with guidance is slightly preferred by human raters for image aesthetics, plausibility and text alignment.

\begin{table}
\resizebox{1.\columnwidth}{!}{
\begin{tabular}{@{}lccccc@{}}
\toprule
Preference & $\gg$ & $>$ & $\approx$ & $<$ & $\ll$\\ \midrule
Percentage & 7.0\% & 20.5\% & 33.2\% & 18.5\% & 4.2\% \\ \bottomrule
\end{tabular}
}
\vspace{-0.5em}
\caption{\textbf{Human Evaluation Results: Latent Diffusion (LD) with overall score as Classifier Guidance vs original Latent Diffusion preference}: 
Percentage of examples where LD with guidance is significantly better ($\gg$), slightly better ($>$), about the same ($\approx$), slightly worse ($<$), significantly worse ($\ll$) than original LD.
Data was collected from 4 individuals in a randomized survey.}
\label{tab:ld_humaneval}
\end{table}
\fi

\end{document}